%% file: main.tex
\documentclass[letterpaper, 10 pt, conference]{ieeeconf}  

\IEEEoverridecommandlockouts                              

\usepackage{graphicx}      
\usepackage{amsmath,amssymb,amsfonts}

\usepackage{comment}
\usepackage[utf8]{inputenc} 
\usepackage{tikz}
\usepackage{algorithmic}
\usepackage{textcomp}
\usepackage{xcolor}
\usepackage{cite}
\usepackage{csquotes}
\usepackage{multicol}
\usepackage[hyphens,spaces,obeyspaces]{url}
\usepackage{mathtools}
\usepackage{xcolor}
\usepackage{siunitx}
\usepackage{hyperref} 
\hypersetup{backref=true,       
    pagebackref=true,               
    hyperindex=true,                
    colorlinks=true,                
    breaklinks=true,                
    urlcolor= black,                
    linkcolor= blue,                
    bookmarks=true,                 
    bookmarksopen=false,
    filecolor=black,
    citecolor=blue,
    linkbordercolor=blue
}

\usepackage{subcaption}
\usepackage{graphicx}
\usepackage{balance}
\usepackage{environ}
\usetikzlibrary{patterns}

\newcommand\numeq[1]%
  {\stackrel{\scriptscriptstyle(\mkern-1.5mu#1\mkern-1.5mu)}{=}}

\newtheorem{assumption}{Assumption}

\usepackage{cancel}

\title{\LARGE \bf
A Control Approach for Human-Robot Ergonomic Payload Lifting
}

\author{Lorenzo Rapetti$^{1,2}$, Carlotta Sartore$^{1,2}$, Mohamed Elobaid$^1$, Yeshasvi Tirupachuri$^1$, \\ Francesco Draicchio$^{3}$, Tomohiro Kawakami$^{4}$,
Takahide Yoshiike$^{4}$, and Daniele Pucci$^{1,2}$
\thanks{$^{1}$Artificial and Mechanical Intelligence at Italian Insititute of Technology, Center for Robotics and Intelligent Systems, Genoa, Italy. (email: {\tt\small { firstname.lastname@iit.it}}).}%
\thanks{$^{2}$Machine Learning and Optimisation, The University of Manchester, Manchester, United Kingdom.}
\thanks{$^{3}$  Department of Occupational and Environmental Medicine, Epidemiology and Hygiene, INAIL, Roma, Italy.}
\thanks{$^{4}$ Honda R\&D Co., Ltd., Saitama, Japan.}
}

\begin{document}

\maketitle
\thispagestyle{empty}
\pagestyle{empty}

\begin{abstract}     
Collaborative robots can relief human operators from excessive efforts during payload lifting activities. Modelling the human partner allows the design of safe and efficient collaborative strategies.
In this paper, we present a control approach for human-robot collaboration based on human monitoring through whole-body wearable sensors, and interaction modelling through coupled rigid-body dynamics. Moreover, a trajectory advancement strategy is proposed, allowing for online adaptation of the robot trajectory depending on the human motion. The resulting framework allows us to perform payload lifting tasks,  taking into
account the ergonomic requirements of the agents. Validation has been performed in an experimental scenario using the iCub3 humanoid robot and a human subject sensorized with the iFeel wearable system.  

\end{abstract}

\smallskip 


\input{sections/introduction}

\input{sections/background}

\input{sections/proposed_architecture}

\input{sections/experimental_validation}

\input{sections/conclusions}

\section*{Acknowledgment}

This work was supported by Honda R\&D Co. Ltd and by the Italian National Institute for Insurance against accidents at Work (INAIL) ergoCub Project.

\balance
\bibliographystyle{IEEEtran}      
\bibliography{biblio}                  

\end{document}

%% file: sections/introduction.tex
\section{Introduction}

In many workplaces, such as warehouses and 
manufacturing industries, payload lifting activities are executed by human operators. Those activities are the main cause of musculoskeletal problems, such as work-related low-back disorders \cite{kuijer2014}, which are the most common and costly musculoskeletal diseases \cite{lu2014}. 
Robots of the future, including humanoid robots, are expected to relief human operators in heavy manual activities, by collaborating with them towards injury risk reduction \cite{vysocky2016,ajoudani2018} without renouncing human dexterity and decision-making. This paper presents a control framework for ergonomic human-robot collaborative lifting.

Human-robot physical interaction has been of great interest to the robotics community since the 1990s. The main strategies are summarized in \cite{desantis2008} and \cite{haddadin2016}, and identify safety and dependability as the principal requirements. Collaboration is a particular case of interaction where both agents, i.e. the human and the robot, pursue a common goal \cite{haddadin2016}. 
While safety remains a prerogative, a fruitful human-robot collaboration depends as well on the agents capability to coordinate the actions and distribute the function to achieve an efficient task execution.
In this perspective, robots should be able to continuously monitor the human collaborator to estimate its state and intentions \cite{bauer2008}.

Humanoid robots resemble human body shapes.
Human-likeness is expected to ease the exploration and interaction with anthropomorphic environments, hence, to facilitate execution of certain collaborative tasks \cite{kajita2014}. Moreover, anthropomorphism seems to positively affect social interaction and empathy toward robots \cite{riek2009}.
In order to maintain humanoid robot balancing and achieve locomotion, \textit{momentum-based} whole-body control strategies have been successfully implemented into different platforms \cite{stephens2010,righetti2011,herzog2014,nava2016,kanemoto2018}, and compliance with respect to external interactions is achieved thanks to motor joint torques control~\cite{hyon2007,pucci2016,ko2018}. 
There are few examples in literature, where whole-body control strategies are used on real humanoid robots to enable payload lifting with a human partner \cite{evrard2009,sheng2014,agravante2019}, but they do not consider the whole dynamics of the two agents during the interaction. In \cite{evrard2009,agravante2019} the human action is modelled through interaction forces, while \cite{sheng2014} propose a kinematics-only strategy. 

In \cite{rapetti2021shared}, we proposed a shared-control framework for robot-robot collaboration allowing two humanoid robots to perform a team lifting task. The solution is based on the usage of a coupled dynamics model, originally proposed in \cite{tirupachuri2020}, which describes the interaction of the two agents with a single set of equations, and on a global optimization for the system dynamics. Moreover, ergonomics principles are introduced and applied for improving efficiency of the task execution.
Starting from those results, we aim at updating the framework by replacing a robot with a sensorized human. 

\begin{figure}[t]
    \centering
        \includegraphics[width=0.42\textwidth]{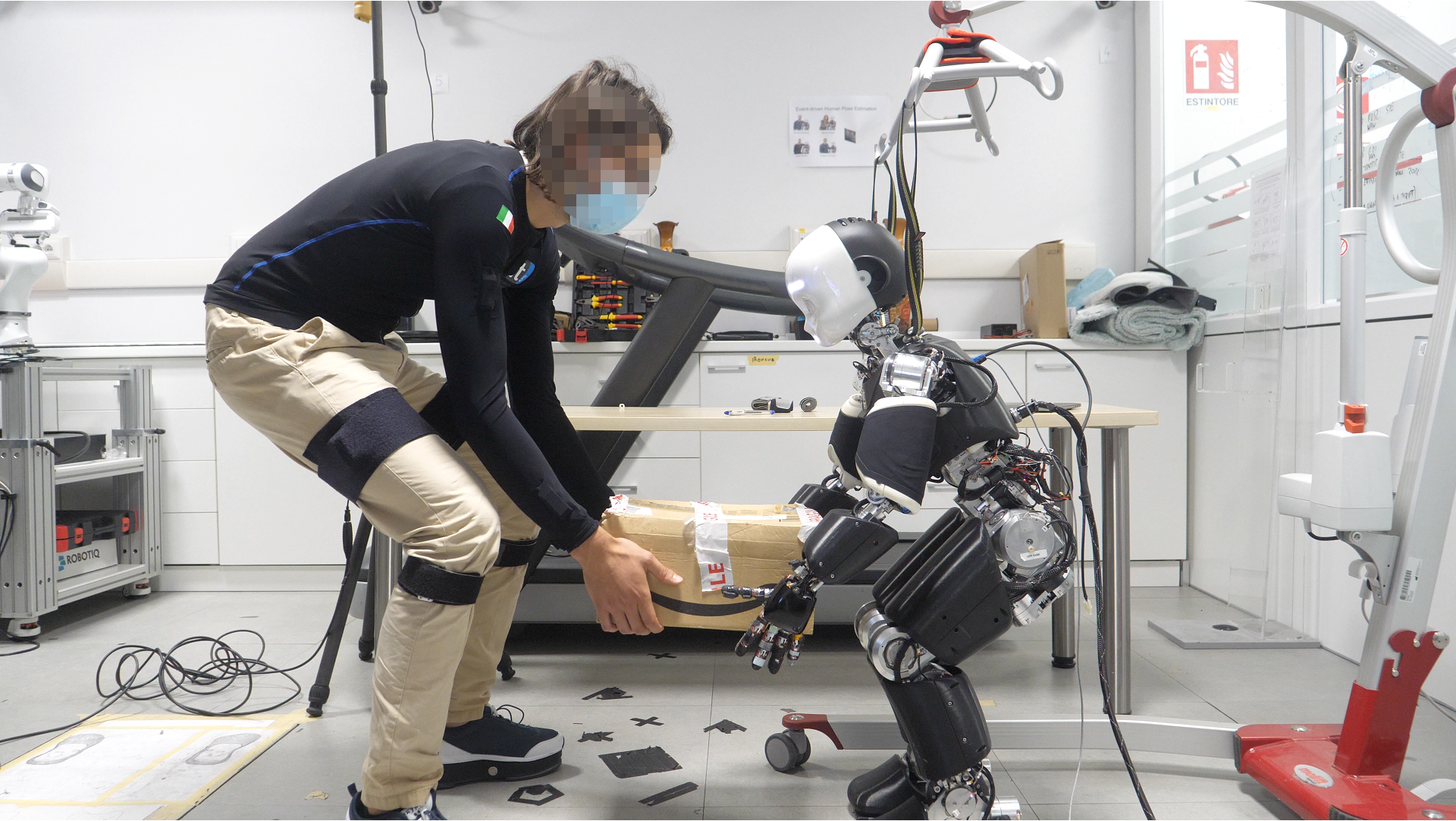}
    \caption[Human-robot collaborative lifting with iCub robot.]{Human-robot collaborative scenario where an iCub humanoid robot and a sensorized human lift a payload.}
        \label{fig:human-robot-collaborative-lifting}
    \end{figure} 

This article presents a control approach that allows humanoid robots to perform collaborative lifting tasks with humans, as shown in Figure \ref{fig:human-robot-collaborative-lifting}. The proposed approach is based on human partner awareness through the estimation of human kinematics and dynamics quantities from wearable sensors data, and it exploits the coupled dynamics model proposed in \cite{rapetti2021shared}. Furthermore, it allows to synchronize agents actions and it is validated on real hardware executing collaborative lifting tasks with the humanoid robot iCub3\cite{dafarra2022} and using the iFeel wearable sensors system.

The manuscript is organized as follows: Section \ref{sec:background} introduces the
notation and modelling. Section \ref{sec:proposed-architecture} presents the proposed control framework. Section \ref{sec:experimental-validation} presents the implementation details and the experimental results. Concluding remarks presented in Section \ref{sec:conclusions} end the manuscript.

%% file: sections/background.tex
\section{Background}
\label{sec:background}

\subsection{Notation}
\label{sec:background:notation}
\begin{itemize}
    \item $\mathcal{I}$ denotes the inertial frame of reference.
    \item  $\prescript{\mathcal{A}}{}{p}_{\mathcal{B}} \in \mathbb{R}^3$ is the the position of the origin of the frame $\mathcal{B}$ with respect to the frame $\mathcal{A}$.
    \item  $\prescript{\mathcal{A}}{}{R}_{\mathcal{B}} \in SO(3)$ represents the rotation matrix of the frame $\mathcal{B}$ with respect to $\mathcal{A}$.
    \item $\prescript{\mathcal{A}}{}{\omega}_{\mathcal{B}} \in \mathbb{R}^3$ is the angular velocity of the frame $\mathcal{B}$ with respect to $\mathcal{A}$, expressed in $\mathcal{A}$.
    \item The operator $\text{sk}(\cdot) :\mathbb{R}^{3 \times 3} \to so(3)$ denotes \textit{skew-symmetric} operation of a matrix, such that given $A \in \mathbb{R}^{3 \times 3}$, it is defined as $\text{sk}(A) := (A - A^\top)/2$.
    \item The \textit{vee} operator $.^{\vee} : so(3) \to \mathbb{R}^{3}$ denotes the inverse of \textit{skew-symmetric} vector operator. Given a matrix $A \in so(3)$ and a vector $u \in \mathbb{R}^{3}$, it is defined as $Au = A^{\vee} \times u$.
    \item $_AX^{B}$ denotes a wrench 6D vector transform, as defined in \cite{traversaro2016multibody}, such that $_AX^{B}=\begin{bmatrix} {}^{A}R_B & 0 \\ S(p_B - p_A) & {}^{A}R_B \end{bmatrix}$.
    \item The operator $\left\lVert . \right\rVert_2$ indicates vector squared norm. Given $v \in \mathbb{R}^{n}$, it is defined as $\left\lVert v \right\rVert_2 = \sqrt{v_1^2+...+v_n^2}$.
    \item The operator $(\cdot)^{\dagger}$ indicates the \textit{Moore-Penrose pseudo-inverse} operator. In case $A$ is a full row rank matrix it can be computed as $A^{\dagger}=A^{T}(A A^{T})^{-1}$.
\end{itemize}

\subsection{Modelling}
\label{sec:background:modelling}
A humanoid robot can be modelled as a multi-body mechanical system composed of rigid \textit{links} connected by \textit{joints}. None of the links has a constant \emph{position-and-orientation} -- equivalently \emph{pose} -- with respect to the inertial frame, and it is said to be \emph{floting-base}. Hence, a specific frame, attached to a link of the system, is referred to as the \textit{base frame} and denoted as $\mathcal{B}$.
The \textit{model configuration} is characterized by the pose of the \textit{base frame} along with the \textit{joint positions}. The configuration space lies on the Lie group $\mathbb{Q}=\mathbb{R}^{3} \times SO(3) \times \mathbb{R}^{n} $. An element of the configuration space $q \in \mathbb{Q}$ is defined as the triplet $q = (\prescript{\mathcal{I}}{}{p}_{\mathcal{B}}, \prescript{\mathcal{I}}{}{R}_{\mathcal{B}}, s)$ where $\prescript{\mathcal{I}}{}{p}_{\mathcal{B}} \in \mathbb{R}^3$ and $\prescript{\mathcal{I}}{}{R}_{\mathcal{B}} \in SO(3)$ denote the position and the orientation of the \textit{base frame} respectively, and $s \in \mathbb{R}^n$ is the joints configuration representing the topology of the mechanical system. The position and orientation of a frame $\mathcal{A}$ can be obtained from the model configuration via a geometrical forward kinematics map $h_{\mathcal{A}}(\cdot):\mathbb{Q} \to SO(3) \times \mathbb{R}^3$.
The \textit{model velocity} is defined by the linear and angular velocity of the \textit{base frame} along with the \textit{joint velocities}. The  velocity space lies on the set $\mathbb{V} = \mathbb{R}^{6+n}$. An element of the velocity space $\nu \in \mathbb{V}$ is defined as $\nu = (\prescript{\mathcal{I}}{}{\mathrm{v}}_{\mathcal{B}}, \dot{s})$ where $\prescript{\mathcal{I}}{}{\mathrm{v}}_{\mathcal{B}}=(\prescript{\mathcal{I}}{}{\dot{p}}_{\mathcal{B}}, \prescript{\mathcal{I}}{}{\omega}_{\mathcal{B}}) \in \mathbb{R}^6$ denotes the linear and angular velocity of the \textit{base frame}, and $\dot{s}$ denotes the joint velocities. The velocity of a frame $\mathcal{A}$ attached to the model is denoted by $\prescript{\mathcal{I}}{}{\mathrm{v}}_{\mathcal{A}}=(\prescript{\mathcal{I}}{}{\dot{p}}_{\mathcal{A}}, \prescript{\mathcal{I}}{}{\omega}_{\mathcal{A}})$ with the linear and the angular velocity components respectively. The mapping between frame velocity $\prescript{\mathcal{I}}{}{\mathrm{v}}_{\mathcal{A}}$ and system velocity $\nu$ is obtained via the \textit{Jacobian} ${J}_{\mathcal{A}}={J}_{\mathcal{A}}(q) \in \mathbb{R}^{6 \times (n+6)}$, i.e. $\prescript{\mathcal{I}}{}{\mathrm{v}}_{\mathcal{A}} = {J}_{\mathcal{A}}(q)  \nu$. 

In a first approximation, a human can be modelled as a multi-body system following the same formalism used for humanoid robots. Hence, throughout this work, we will refer to the human model as a simplified model composed of simple geometrical shapes. This choice allows for fast processing, while trying to capture the dominant dynamics.

\subsection{Coupled Dynamics}
\label{sec:coupled-dynamics}
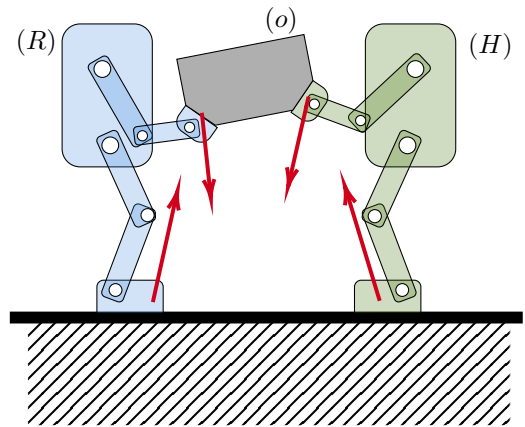
\begin{figure}[b]
      \centering
      \tikzset{every picture/.style={scale=0.65,every picture/.style={}}}
      \input{Figs/collaborative-models.tikz}
      \caption{Example of physical interaction between two multi rigid-body models, representing a robot and a human, in contact through an object.}
      \label{fig:model-contacts}
  \end{figure}
  
The dynamics of each agent is described applying the Euler-Poincarè formalism \cite{Marsden2010}.
 Agent dynamics is thus described by a set of differential equations complemented with holonomic constraints characterising the contacts:
\begin{align}
\label{eq:constrained-dynamic}
\begin{split}
    & M(q) \dot{\nu} + h(q,\nu) = B {\tau} + J_c^T \mathrm{f}, \\
    & J_c \ \nu = 0,
\end{split}
\end{align}
where $M \in \mathbb{R}^{n+6 \times n+6}$ is the mass matrix, the term $h \in \mathbb{R}^{n+6}$ accounts for Coriolis and gravity forces, $B = (0_{n \times 6},I_n)^T$ is a selector matrix, ${\tau}  \in \mathbb{R}^{n}$ is a vector representing the agent joint torques, $\mathrm{f} \in \mathbb{R}^{6n_c}$ represents the wrenches acting on $n_c$ contact links of the agent, and $J_c \in \mathbb{R}^{n+6 \times 6n_c}$ is the contact Jacobian. Now, consider the case shown in Figure \ref{fig:model-contacts} where a robot, denoted with the subscript $\text{(R)}$, and a human, denoted with the subscript $\text{(H)}$, are physically interacting through an object, denote with the subscript $(o)$. The contacts are not limited to those with the environment, but agent-object-agent contacts should also be considered. Denoting composite matrices with $\mathbf{bold}$ font, the coupled system dynamics is described as follow \cite{tirupachuri2020}
\begin{subequations}
\label{eq:multi-system-equations-compact}
\begin{align}
\label{eq:multi-system-equations-compact:dynamics}
& \mathbf{M} \dot{\boldsymbol{\nu}} + \mathbf{h} = \mathbf{B} \boldsymbol{\tau} + \mathbf{Q}^T \mathbf{f}, \\
\label{eq:multi-system-equations-compact:holonomics}
& \mathbf{Q} \boldsymbol{\nu} = 0.
\end{align}
\end{subequations}
The structure of matrices $\mathbf{M}$, $\boldsymbol{\nu}$, $\mathbf{h}$, $\mathbf{B}$, $\boldsymbol{\tau}$, $\mathbf{Q}$, and $\mathbf{f}$ is presented in \cite{tirupachuri2020} and \cite{rapetti2021shared}.


%% file: Figs/collaborative-models.tikz
 
\tikzset{
pattern size/.store in=\mcSize, 
pattern size = 5pt,
pattern thickness/.store in=\mcThickness, 
pattern thickness = 0.3pt,
pattern radius/.store in=\mcRadius, 
pattern radius = 1pt}
\makeatletter
\pgfutil@ifundefined{pgf@pattern@name@_p7pb65jkf}{
\pgfdeclarepatternformonly[\mcThickness,\mcSize]{_p7pb65jkf}
{\pgfqpoint{0pt}{0pt}}
{\pgfpoint{\mcSize+\mcThickness}{\mcSize+\mcThickness}}
{\pgfpoint{\mcSize}{\mcSize}}
{
\pgfsetcolor{\tikz@pattern@color}
\pgfsetlinewidth{\mcThickness}
\pgfpathmoveto{\pgfqpoint{0pt}{0pt}}
\pgfpathlineto{\pgfpoint{\mcSize+\mcThickness}{\mcSize+\mcThickness}}
\pgfusepath{stroke}
}}
\makeatother

\begin{tikzpicture}[x=0.75pt,y=0.75pt,yscale=-1,xscale=1]

\draw  [color={rgb, 255:red, 255; green, 255; blue, 255 }  ,draw opacity=1 ][pattern=_p7pb65jkf,pattern size=5.475pt,pattern thickness=0.75pt,pattern radius=0pt, pattern color={rgb, 255:red, 0; green, 0; blue, 0}] (162.23,406.15) -- (536.73,406.15) -- (536.73,488.59) -- (162.23,488.59) -- cycle ;
\draw [line width=4.5]    (148.43,404.59) -- (546.43,404.59) ;
\draw  [fill={rgb, 255:red, 74; green, 144; blue, 226 }  ,fill opacity=0.25 ] (216.01,380.77) .. controls (216.01,377.9) and (218.34,375.58) .. (221.21,375.58) -- (262.24,375.61) .. controls (265.11,375.61) and (267.44,377.94) .. (267.43,380.81) -- (267.42,401.61) .. controls (267.42,401.61) and (267.42,401.61) .. (267.42,401.61) -- (215.99,401.57) .. controls (215.99,401.57) and (215.99,401.57) .. (215.99,401.57) -- cycle ;
\draw  [fill={rgb, 255:red, 74; green, 144; blue, 226 }  ,fill opacity=0.25 ] (189,190.59) .. controls (189,182.86) and (195.27,176.59) .. (203,176.59) -- (245,176.59) .. controls (252.73,176.59) and (259,182.86) .. (259,190.59) -- (259,273.59) .. controls (259,281.32) and (252.73,287.59) .. (245,287.59) -- (203,287.59) .. controls (195.27,287.59) and (189,281.32) .. (189,273.59) -- cycle ;
\draw  [fill={rgb, 255:red, 74; green, 144; blue, 226 }  ,fill opacity=0.25 ] (222.03,388.48) .. controls (220.19,387.79) and (219.27,385.74) .. (219.96,383.91) -- (244.58,318.65) .. controls (245.27,316.82) and (247.32,315.89) .. (249.15,316.58) -- (259.11,320.34) .. controls (260.94,321.03) and (261.87,323.08) .. (261.18,324.91) -- (236.56,390.17) .. controls (235.86,392.01) and (233.82,392.93) .. (231.98,392.24) -- cycle ;
\draw  [fill={rgb, 255:red, 74; green, 144; blue, 226 }  ,fill opacity=0.25 ] (249.71,334) .. controls (247.93,334.81) and (245.83,334.02) .. (245.02,332.23) -- (216.24,268.69) .. controls (215.43,266.91) and (216.22,264.81) .. (218,264) -- (227.7,259.61) .. controls (229.48,258.8) and (231.59,259.59) .. (232.4,261.38) -- (261.18,324.91) .. controls (261.98,326.7) and (261.19,328.8) .. (259.41,329.61) -- cycle ;
\draw  [fill={rgb, 255:red, 74; green, 144; blue, 226 }  ,fill opacity=0.25 ] (248.15,271.57) .. controls (246.46,272.55) and (244.3,271.97) .. (243.32,270.28) -- (209.74,212.06) .. controls (208.76,210.37) and (209.34,208.21) .. (211.03,207.23) -- (220.14,201.98) .. controls (221.83,201) and (223.99,201.58) .. (224.97,203.27) -- (258.55,261.49) .. controls (259.53,263.18) and (258.95,265.34) .. (257.26,266.32) -- cycle ;
\draw  [fill={rgb, 255:red, 74; green, 144; blue, 226 }  ,fill opacity=0.25 ] (295.81,261.51) .. controls (296.01,263.07) and (294.91,264.51) .. (293.35,264.71) -- (248.79,270.57) .. controls (247.22,270.77) and (245.78,269.67) .. (245.58,268.1) -- (244.46,259.59) .. controls (244.25,258.03) and (245.36,256.59) .. (246.92,256.38) -- (291.48,250.53) .. controls (293.05,250.33) and (294.49,251.43) .. (294.69,253) -- cycle ;
\draw  [fill={rgb, 255:red, 74; green, 144; blue, 226 }  ,fill opacity=0.25 ] (304.12,264.53) .. controls (301.19,268.66) and (295.48,269.64) .. (291.35,266.71) -- (281.86,260) .. controls (277.73,257.08) and (276.75,251.36) .. (279.67,247.23) -- (284.96,239.75) .. controls (284.96,239.75) and (284.96,239.75) .. (284.96,239.75) -- (309.41,257.05) .. controls (309.41,257.05) and (309.41,257.05) .. (309.41,257.05) -- cycle ;
\draw  [fill={rgb, 255:red, 255; green, 255; blue, 255 }  ,fill opacity=1 ] (213.73,211.62) .. controls (213.73,208.03) and (216.64,205.12) .. (220.23,205.12) .. controls (223.82,205.12) and (226.73,208.03) .. (226.73,211.62) .. controls (226.73,215.21) and (223.82,218.12) .. (220.23,218.12) .. controls (216.64,218.12) and (213.73,215.21) .. (213.73,211.62) -- cycle ;
\draw  [fill={rgb, 255:red, 255; green, 255; blue, 255 }  ,fill opacity=1 ] (219.9,270.88) .. controls (219.9,267.29) and (222.81,264.38) .. (226.4,264.38) .. controls (229.99,264.38) and (232.9,267.29) .. (232.9,270.88) .. controls (232.9,274.47) and (229.99,277.38) .. (226.4,277.38) .. controls (222.81,277.38) and (219.9,274.47) .. (219.9,270.88) -- cycle ;
\draw  [fill={rgb, 255:red, 255; green, 255; blue, 255 }  ,fill opacity=1 ] (250.21,325.11) .. controls (250.21,322.29) and (252.5,320) .. (255.32,320) .. controls (258.14,320) and (260.43,322.29) .. (260.43,325.11) .. controls (260.43,327.93) and (258.14,330.21) .. (255.32,330.21) .. controls (252.5,330.21) and (250.21,327.93) .. (250.21,325.11) -- cycle ;
\draw  [fill={rgb, 255:red, 255; green, 255; blue, 255 }  ,fill opacity=1 ] (225.48,383.21) .. controls (225.48,380.47) and (227.71,378.24) .. (230.46,378.24) .. controls (233.2,378.24) and (235.43,380.47) .. (235.43,383.21) .. controls (235.43,385.96) and (233.2,388.18) .. (230.46,388.18) .. controls (227.71,388.18) and (225.48,385.96) .. (225.48,383.21) -- cycle ;
\draw  [fill={rgb, 255:red, 255; green, 255; blue, 255 }  ,fill opacity=1 ] (247.38,263.43) .. controls (247.38,261.25) and (249.14,259.48) .. (251.32,259.48) .. controls (253.5,259.48) and (255.27,261.25) .. (255.27,263.43) .. controls (255.27,265.61) and (253.5,267.37) .. (251.32,267.37) .. controls (249.14,267.37) and (247.38,265.61) .. (247.38,263.43) -- cycle ;
\draw  [fill={rgb, 255:red, 255; green, 255; blue, 255 }  ,fill opacity=1 ] (284,256.88) .. controls (284,254.7) and (285.77,252.93) .. (287.95,252.93) .. controls (290.13,252.93) and (291.89,254.7) .. (291.89,256.88) .. controls (291.89,259.06) and (290.13,260.83) .. (287.95,260.83) .. controls (285.77,260.83) and (284,259.06) .. (284,256.88) -- cycle ;
\draw  [fill={rgb, 255:red, 65; green, 117; blue, 5 }  ,fill opacity=0.25 ] (467.38,380.77) .. controls (467.38,377.9) and (465.05,375.58) .. (462.18,375.58) -- (421.06,375.61) .. controls (418.19,375.61) and (415.87,377.94) .. (415.87,380.81) -- (415.88,401.61) .. controls (415.88,401.61) and (415.88,401.61) .. (415.88,401.61) -- (467.39,401.57) .. controls (467.39,401.57) and (467.39,401.57) .. (467.39,401.57) -- cycle ;
\draw  [fill={rgb, 255:red, 65; green, 117; blue, 5 }  ,fill opacity=0.25 ] (494.43,190.61) .. controls (494.43,182.87) and (488.15,176.59) .. (480.41,176.59) -- (438.34,176.59) .. controls (430.59,176.59) and (424.32,182.87) .. (424.32,190.61) -- (424.32,273.57) .. controls (424.32,281.31) and (430.59,287.59) .. (438.34,287.59) -- (480.41,287.59) .. controls (488.15,287.59) and (494.43,281.31) .. (494.43,273.57) -- cycle ;
\draw  [fill={rgb, 255:red, 65; green, 117; blue, 5 }  ,fill opacity=0.25 ] (461.35,388.48) .. controls (463.19,387.79) and (464.11,385.74) .. (463.42,383.91) -- (438.76,318.65) .. controls (438.07,316.82) and (436.02,315.89) .. (434.18,316.58) -- (424.21,320.34) .. controls (422.37,321.03) and (421.44,323.08) .. (422.14,324.92) -- (446.8,390.17) .. controls (447.49,392) and (449.54,392.93) .. (451.38,392.24) -- cycle ;
\draw  [fill={rgb, 255:red, 65; green, 117; blue, 5 }  ,fill opacity=0.25 ] (433.61,333.99) .. controls (435.4,334.8) and (437.51,334.01) .. (438.32,332.22) -- (467.15,268.7) .. controls (467.96,266.91) and (467.17,264.81) .. (465.38,264) -- (455.67,259.61) .. controls (453.88,258.8) and (451.78,259.6) .. (450.97,261.38) -- (422.14,324.91) .. controls (421.33,326.7) and (422.12,328.8) .. (423.91,329.61) -- cycle ;
\draw  [fill={rgb, 255:red, 65; green, 117; blue, 5 }  ,fill opacity=0.25 ] (419.36,259.24) .. controls (420.69,260.67) and (422.93,260.74) .. (424.36,259.41) -- (473.62,213.56) .. controls (475.05,212.23) and (475.13,209.99) .. (473.8,208.57) -- (466.62,200.87) .. controls (465.28,199.45) and (463.05,199.37) .. (461.62,200.7) -- (412.36,246.56) .. controls (410.93,247.89) and (410.85,250.12) .. (412.18,251.55) -- cycle ;
\draw  [fill={rgb, 255:red, 65; green, 117; blue, 5 }  ,fill opacity=0.25 ] (377.22,241.16) .. controls (376.67,242.64) and (377.42,244.29) .. (378.91,244.84) -- (421.08,260.5) .. controls (422.56,261.05) and (424.21,260.29) .. (424.77,258.81) -- (427.77,250.76) .. controls (428.32,249.28) and (427.57,247.63) .. (426.09,247.08) -- (383.91,231.42) .. controls (382.43,230.87) and (380.78,231.63) .. (380.23,233.11) -- cycle ;
\draw  [fill={rgb, 255:red, 65; green, 117; blue, 5 }  ,fill opacity=0.25 ] (374.23,250.44) .. controls (378.41,253.31) and (384.12,252.25) .. (386.99,248.07) -- (393.55,238.5) .. controls (396.42,234.32) and (395.35,228.61) .. (391.17,225.75) -- (383.6,220.56) .. controls (383.6,220.56) and (383.6,220.56) .. (383.6,220.56) -- (366.66,245.26) .. controls (366.66,245.26) and (366.66,245.26) .. (366.66,245.26) -- cycle ;
\draw  [fill={rgb, 255:red, 255; green, 255; blue, 255 }  ,fill opacity=1 ] (469.66,211.62) .. controls (469.66,208.03) and (466.75,205.12) .. (463.15,205.12) .. controls (459.56,205.12) and (456.64,208.03) .. (456.64,211.62) .. controls (456.64,215.21) and (459.56,218.12) .. (463.15,218.12) .. controls (466.75,218.12) and (469.66,215.21) .. (469.66,211.62) -- cycle ;
\draw  [fill={rgb, 255:red, 255; green, 255; blue, 255 }  ,fill opacity=1 ] (463.48,270.88) .. controls (463.48,267.29) and (460.57,264.38) .. (456.97,264.38) .. controls (453.38,264.38) and (450.46,267.29) .. (450.46,270.88) .. controls (450.46,274.47) and (453.38,277.38) .. (456.97,277.38) .. controls (460.57,277.38) and (463.48,274.47) .. (463.48,270.88) -- cycle ;
\draw  [fill={rgb, 255:red, 255; green, 255; blue, 255 }  ,fill opacity=1 ] (436.73,325.89) .. controls (436.73,323.07) and (434.44,320.78) .. (431.61,320.78) .. controls (428.79,320.78) and (426.5,323.07) .. (426.5,325.89) .. controls (426.5,328.71) and (428.79,330.99) .. (431.61,330.99) .. controls (434.44,330.99) and (436.73,328.71) .. (436.73,325.89) -- cycle ;
\draw  [fill={rgb, 255:red, 255; green, 255; blue, 255 }  ,fill opacity=1 ] (457.89,383.21) .. controls (457.89,380.47) and (455.66,378.24) .. (452.91,378.24) .. controls (450.16,378.24) and (447.93,380.47) .. (447.93,383.21) .. controls (447.93,385.96) and (450.16,388.18) .. (452.91,388.18) .. controls (455.66,388.18) and (457.89,385.96) .. (457.89,383.21) -- cycle ;
\draw  [fill={rgb, 255:red, 255; green, 255; blue, 255 }  ,fill opacity=1 ] (424.59,251.64) .. controls (424.59,249.46) and (422.82,247.69) .. (420.63,247.69) .. controls (418.45,247.69) and (416.68,249.46) .. (416.68,251.64) .. controls (416.68,253.82) and (418.45,255.59) .. (420.63,255.59) .. controls (422.82,255.59) and (424.59,253.82) .. (424.59,251.64) -- cycle ;
\draw  [fill={rgb, 255:red, 255; green, 255; blue, 255 }  ,fill opacity=1 ] (388.69,238.87) .. controls (388.69,236.69) and (386.92,234.93) .. (384.74,234.93) .. controls (382.56,234.93) and (380.79,236.69) .. (380.79,238.87) .. controls (380.79,241.05) and (382.56,242.82) .. (384.74,242.82) .. controls (386.92,242.82) and (388.69,241.05) .. (388.69,238.87) -- cycle ;
\draw [color={rgb, 255:red, 208; green, 2; blue, 27 }  ,draw opacity=1 ][fill={rgb, 255:red, 208; green, 2; blue, 27 }  ,fill opacity=1 ][line width=1.5]    (259.43,392.59) -- (277.78,309.52) ;
\draw [shift={(278.43,306.59)}, rotate = 102.46] [color={rgb, 255:red, 208; green, 2; blue, 27 }  ,draw opacity=1 ][line width=1.5]    (14.21,-4.28) .. controls (9.04,-1.82) and (4.3,-0.39) .. (0,0) .. controls (4.3,0.39) and (9.04,1.82) .. (14.21,4.28)   ;
\draw [color={rgb, 255:red, 208; green, 2; blue, 27 }  ,draw opacity=1 ][fill={rgb, 255:red, 208; green, 2; blue, 27 }  ,fill opacity=1 ][line width=1.5]    (435.43,391.59) -- (409.3,305.46) ;
\draw [shift={(408.43,302.59)}, rotate = 73.12] [color={rgb, 255:red, 208; green, 2; blue, 27 }  ,draw opacity=1 ][line width=1.5]    (14.21,-4.28) .. controls (9.04,-1.82) and (4.3,-0.39) .. (0,0) .. controls (4.3,0.39) and (9.04,1.82) .. (14.21,4.28)   ;
\draw  [fill={rgb, 255:red, 155; green, 155; blue, 155 }  ,fill opacity=0.8 ] (284.96,239.75) -- (307.95,255.24) -- (368.11,243.54) -- (383.6,220.56) -- (376.65,184.83) -- (376.65,184.83) -- (278.01,204.02) -- (278.01,204.02) -- cycle ;
\draw [color={rgb, 255:red, 208; green, 2; blue, 27 }  ,draw opacity=1 ][fill={rgb, 255:red, 208; green, 2; blue, 27 }  ,fill opacity=1 ][line width=1.5]    (297.43,245.59) -- (304.11,308.73) ;
\draw [shift={(304.43,311.71)}, rotate = 263.96] [color={rgb, 255:red, 208; green, 2; blue, 27 }  ,draw opacity=1 ][line width=1.5]    (14.21,-4.28) .. controls (9.04,-1.82) and (4.3,-0.39) .. (0,0) .. controls (4.3,0.39) and (9.04,1.82) .. (14.21,4.28)   ;
\draw [color={rgb, 255:red, 208; green, 2; blue, 27 }  ,draw opacity=1 ][fill={rgb, 255:red, 208; green, 2; blue, 27 }  ,fill opacity=1 ][line width=1.5]    (365.09,299.79) -- (380.23,233.11) ;
\draw [shift={(364.43,302.71)}, rotate = 282.79] [color={rgb, 255:red, 208; green, 2; blue, 27 }  ,draw opacity=1 ][line width=1.5]    (14.21,-4.28) .. controls (9.04,-1.82) and (4.3,-0.39) .. (0,0) .. controls (4.3,0.39) and (9.04,1.82) .. (14.21,4.28)   ;

\draw (151.12,178.53) node [anchor=north west][inner sep=0.75pt]   [align=left] {$\displaystyle ( R)$};
\draw (502.12,181.53) node [anchor=north west][inner sep=0.75pt]   [align=left] {$\displaystyle ( H)$};
\draw (344.12,161.53) node [anchor=north west][inner sep=0.75pt]   [align=left] {$\displaystyle ( o)$};

\end{tikzpicture}

%% file: sections/proposed_architecture.tex
  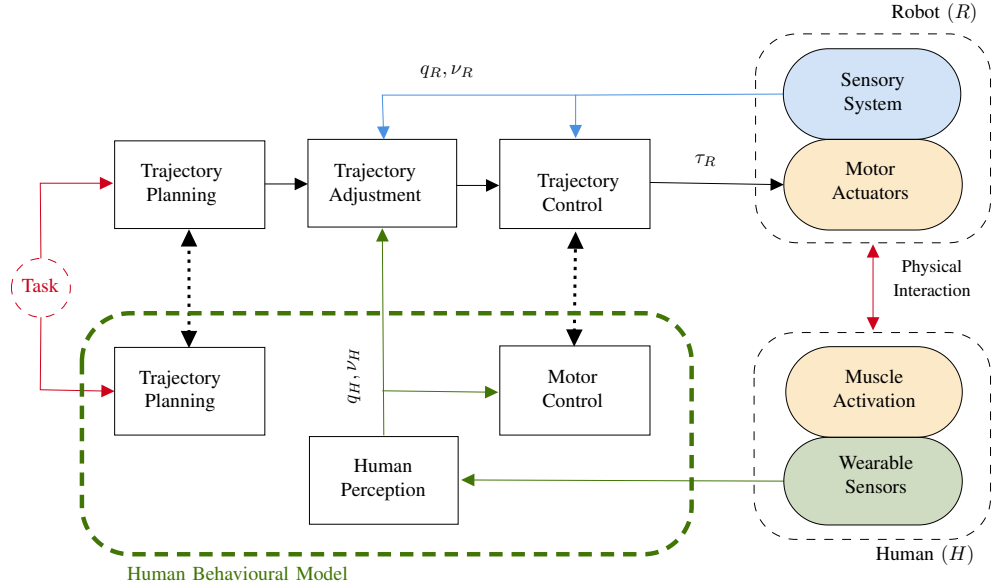
\begin{figure*}[t]
      \centering
      \scalebox{0.75}{
      \tikzset{every picture/.style={scale=1,every picture/.style={}}}
      \input{Figs/control_architecture.tikz}
      }
      \caption[Collaborative control framework for human-robot collaboration.]{Ergonomics control framework for human-robot collaboration. }
      \label{fig:ergonomy-control-framwork}
  \end{figure*}
  
\section{Proposed Approach}
\label{sec:proposed-architecture}

The proposed architecture for human-robot collaborative payload manipulation is shown in Figure \ref{fig:ergonomy-control-framwork}. The robot is controlled via a hierarchical control framework, while the human is monitored trough a wearable sensors system. The characterizing feature of the framework is whole-body partner feedback. The interconnection between the blocks is not limited to data flow (solid lines) but it assumes reciprocal model awareness (dashed lines).
In this Section, the main blocks composing the architecture are discussed.

  \subsection{Human Behavioural Model}
  The collaborative controller embeds a representation of the human partner behaviour depending on its state, on the environment, and on the task that has to be achieved. The representation is divided into three parts: \textit{human perception}, \textit{trajectory planning},   and \textit{motor control}. 
  
  \subsubsection{Human Perception}
  The human perception block retrieves online an estimate of the human configuration from sensors measurement. 
  The model kinematics, i.e. the model configuration $q_{(\text{H})}$ and velocity $\nu_{(\text{H})}$, are retrieved online from inertial measurements using dynamical inverse kinematics algorithms \cite{rapetti2020}. Contact wrenches exchanged with the ground are measured using force-torque sensorized shoes.
  
  \subsubsection{Human Trajectory Planning}
  The trajectory planning block models the motion to be executed by the human to achieve the specific task. 
  In order to accomplish the collaborative task, the motion of both human and robot are mutually dependent, and should be planned considering the presence of the other agent. Hence, in the proposed approach, robot and human trajectories are planned within a single optimization program.

  \subsubsection{Human Motor Control}
  The motor control block models the response of the human to external stimuli. 
  In the proposed approach, the momentum-control model, generally applied to humanoid robots, is used as reference model for the human
  \begin{subequations}
  \label{eq:momentum-control-human}
  \begin{align}
      \dot{{L}}_{(\text{H})}^*  : & = \dot{{L}}_{(\text{H})}^{\text{des}}-K^{(\text{H})}_d({L}_{(\text{H})}-{L}_{(\text{H})}^{\text{des}}), \label{eq:object-control-human-robot} \\
      \dot{L}_{(\text{H})} & = {}_{\mathcal{G}_{(H)}}X^{(\text{H})} \cdot \mathrm{f}_{(\text{H})}+m_{(\text{H})} \mathfrak{g}, 
      \label{eq:object-momentum}
  \end{align}
  \end{subequations}
  where ${{L}}_{(\text{H})} \in \mathbb{R}^6$ is the human linear and angular momentum, $\mathcal{G}_{(H)}$ is the human center of mass, and $K^{(\text{H})}_d \in \mathbb{R}^{6 \times 6}$ is a symmetric positive definite matrix.
  Finding a set of gains consistent with the human behaviour is not trivial, for this reason further simplifications are discussed in Section \ref{sec:experimental-validation}. 
  In other words, we are assuming that the human generate muscular forces in a way similar to how humanoid robots generate joint torques.

  \subsection{Trajectory Planning and Advancement}
  \label{sec:trajectory-planning-and-advancement}
  The trajectory planning is achieved applying the postural ergonomics optimization introduced in \cite{rapetti2021shared}. In a nutshell, the postures of the human and the robot model are obtained from a non-linear optimization problem that constraints the payload position and tries to minimize the joint torques of the two agents for the static case. As an outcome, $n_{\text{ref}}$ reference robot postures $q^{\text{ref}}_{(\text{R}),0}, \hdots, q^{\text{ref}}_{(\text{R}),n_{\text{ref}}}$ are obtained. 

  The trajectory advancement block has the role to regulate the transition between reference robot postures, synchronizing the robot motion with the human. This is achieved by introducing a time-dependent \textit{free parameter} $\psi \in [0, n_{\text{ref}}], \psi = \psi(t)$, following the idea previously presented in \cite{tirupachuri2019}.

  \begin{assumption}
  \label{asmp:fixed-frame}
  There exists a frame $\mathcal{B}$ in the model, that has a constant reference pose in the desired configuration, i.e. ${}^{\mathcal{I}}H^{\text{ref}}_{\mathcal{B},i}={}^{\mathcal{I}}\bar{H}^{\text{ref}}_{\mathcal{B}} \ \forall i=1,\hdots,n_{\text{ref}}$. The frame $\mathcal{B}$ is assumed to be the base frame of the model.
  \end{assumption}

  The Assumption \ref{asmp:fixed-frame} is valid when a foot remains always in contact with the ground. Under this assumption, the transition between the reference postures depends exclusively on the joints configuration. A piece-wise linear model defines the transition between the joint configurations:
  \begin{equation}
      s^{\text{des}}_{(\text{R})}(t) = \begin{cases}
      s^{\text{ref}}_{(\text{R}),0} + \Delta s_{(\text{R}),0} \cdot \psi(t) & 0 \leq \psi(t) < 1, \\ 
      \vdots &  \\
      s^{\text{ref}}_{(\text{R}),i} + \Delta s_{(\text{R}),i} \cdot (\psi(t) - i - 1) & i \leq \psi(t) < i+1,
      \end{cases}
  \end{equation}
  where $\Delta s_{(\text{R}),i}=s^{\text{ref}}_{(\text{R}),i+1} - s^{\text{ref}}_{(\text{R}),i}$. The right derivative is used to approximate $\dot{s}(t)$, such that
  \begin{equation}
  \label{eq:approximate-derivative-joint}
  \begin{aligned}
      \tilde{\dot{s}}^{\text{des}}(t) \approx \dot{s}^{\text{ref}}_{+}(t) & = \begin{cases}
      \Delta s_{(\text{R}),1} \cdot \dot{\psi}(t) & 0 \leq \psi(t) < 1, \\ 
      \vdots &  \\
      \Delta s_{(\text{R}),i} \cdot \dot{\psi}(t) & i \leq \psi(t) < i+1,
      \end{cases} \\
      & = \Delta s(\psi(t)) \cdot \dot{\psi}(t) ,
  \end{aligned}
  \end{equation}
  where $ \Delta s(\psi(t)) \in \mathbb{R}^{n_j \times n_j}$ is a diagonal matrix where $\Delta s_{j,j}=\Delta s_{(\text{R}),j}$.
  In order to regulate the dynamics of the free-variable, a set of task-space targets is defined starting from human measurements. For the sake of simplicity, we consider the case in which the position of the robot hand frame $\mathcal{T}_{(\text{R})}$, should follow the human hand frame $\mathcal{T}_{(\text{H})}$, i.e. ${}^{\mathcal{I}}p_{\mathcal{T}_{(\text{R})}}^{\text{des}}(t) = {}^{\mathcal{I}}p_{\mathcal{T}_{(\text{H})}}(t) + {\text{offset}}$ and ${}^{\mathcal{I}}\dot{p}_{\mathcal{T}_{(\text{R})}}^{\text{des}}(t) = {}^{\mathcal{I}}\dot{p}_{\mathcal{T}_{(\text{H})}}(t)$. Hence, the following desired dynamics is defined
  \begin{subequations}
  \label{eq:single-target-pose-tracking}
  \begin{align}
      {}^{\mathcal{I}}\dot{p}_{\mathcal{T}_{(\text{R})}}^{*}(t) : & = {}^{\mathcal{I}}\dot{p}_{\mathcal{T}_{(\text{R})}}^{\text{des}}(t) + K_{\text{track}}  \left (
        {}^{\mathcal{I}}p_{\mathcal{T}_{(\text{R})}}^{\text{des}}(t) - {}^{\mathcal{I}}p_{\mathcal{T}_{(\text{R})}}(t) \right ), \\
      \label{eq:jacobian-for-pose-tracking}
      {}^{\mathcal{I}}\dot{p}_{\mathcal{T}_{(\text{R})}}^{*}(t) & = J_{\mathcal{T}_{(\text{R})}} \nu(t),
  \end{align}
  \end{subequations}
  where $K_{\text{track}} \in \mathbb{R}^{6 \times 6}$ is a positive definite gain matrix. From Assumption \ref{asmp:fixed-frame} it holds that $\mathrm{v}^{\text{ref}}_{\mathcal{B},(\text{R})}=0$, and using the approximate derivative of $\dot{s}$ in \eqref{eq:approximate-derivative-joint} we can write Eq. \eqref{eq:jacobian-for-pose-tracking} as
  \begin{equation}
      {}^{\mathcal{I}}\dot{p}_{\mathcal{T}}^{*}(t) = J^s_{\mathcal{T}_{(\text{R})}} \cdot \Delta s(\psi(t)) \cdot \dot{\psi}(t),
  \end{equation}
  from which, given $\psi(0)=0$, we have:
  \begin{equation}
      \begin{cases}
        \psi(t_k)=\psi(t_{k-1}) + \dot{\psi}(t_k) \cdot \Delta t, \\
        \dot{\psi}(t_k) = (J^s_{\mathcal{T}_{(\text{R})}}(t_{k-1}) \cdot \Delta s (t_{k-1}) )^{\dagger} \cdot {}^{\mathcal{I}}\dot{p}_{\mathcal{T}_{(\text{R})}}^{*}(t)(t_k).
      \end{cases}
  \end{equation}
 
  \subsection{Motor Control}
  \label{sec:motor-control-robot}
  The robot trajectory control block takes into consideration the human motor control model, and coordinates the robot balancing and the task execution, defining the desired robot joint torques. In order to facilitate the understanding, trajectory control can be divided in two parts, namely the \textit{force optimization} and the \textit{task-space inverse coupled dynamics}. As shown later in Section \ref{sec:experimental_validatoin_implementation}, the two parts are implemented as a single optimization that can be solved as Quadratic Programming (QP) optimization.
  
  \subsubsection{Force Optimization}
  The force optimization is designed as a set of constraints on the contact wrenches for the achievement of the following tasks:
  \begin{itemize}
      \item control of the robot momentum
      \begin{subequations}
      \label{eq:momentum-control}
       \begin{align}
       \label{eq:momentum-control-i}
      \dot{{L}}_{(\text{R})}^*  : & = \dot{{L}}_{(\text{R})}^{\text{des}}-K^{(\text{R})}_d({L}_{(\text{R})}-{L}_{(\text{R})}^{\text{des}}), \\
      \label{eq:momentum-dynamics-i}
      \dot{L}_{(\text{R})}(\mathrm{f}_{(\text{R})}) & = {}_{\mathcal{G}_{(\text{R})}}X^{(\text{R})} \cdot \mathrm{f}_{(\text{R})}+m_{(\text{R})} \mathfrak{g},
      \end{align}
      \end{subequations}
      where ${{L}}_{(\text{R})} \in \mathbb{R}^6$ is the robot linear and angular momentum, and $K^{(\text{R})}_d \in \mathbb{R}^{6 \times 6}$ is a symmetric positive definite weight matrix;
      \item control of the payload position and orientation \cite{olfatti-saber2001}
      \begin{subequations}
     \label{eq:object-pose-control}
     \begin{align}
    \dot{\mathrm{v}}_{(o)}^* := & \begin{bmatrix} \ddot{p}_{(o)}^{\text{des}} \\\ \dot{\omega}_{(o)}^{\text{des}} \end{bmatrix}-K^{(o)}_d \begin{bmatrix} \dot{p}_{(o)}-\dot{p}_{(o)}^{\text{des}} \\\ \omega_{(o)} - \omega_{(o)}^{\text{des}} \end{bmatrix} \notag \\ 
    & - K^{(o)}_p  \begin{bmatrix}  p_{(o)}-p_{(o)}^{\text{des}} \\\ \text{sk}(R_{(o)}{R_{(o)}^{\text{des}}}^T)^{\vee}
    \end{bmatrix}, \\
    \label{eq:rigid-body-dynamics-compact}
     M_{(o)}\dot{\mathrm{v}}_{(o)}& ={}_{\mathcal{G}_o}X^{(o)}\mathrm{f}_{(o)}+m_{(o)}\mathfrak{g},
    \end{align}
    \end{subequations}
     where $K^{(o)}_d \in \mathbb{R}^{6\times6}$ and $K^{(o)}_p \in \mathbb{R}^{6\times6}$ are positive definite diagonal gains matrices, and the location of the contacts is assumed to be known, as well as the object pose, \textit{e.g.} via vision;
    \item control of the human momentum accordingly to the human motor control model in Equation \eqref{eq:momentum-control-human}.
    \end{itemize}
  The tasks in Equation \eqref{eq:momentum-control-human}, \eqref{eq:momentum-control} and \eqref{eq:object-pose-control},  are linear with respect to the contact wrenches, and can be expressed as
  \begin{equation}
  \label{eq:compact-force-tasks}
      \mathbf{b}^{*} = \mathbf{A} \cdot \mathbf{f}.
  \end{equation}
  
  In order to ensure feasible contacts, additional constraints should be added on the contact wrenches in order to grant unilaterality of contacts, static friction constraints, and ensuring the center of pressure is contained within the contact surface.
  As shown in \cite{caron2015}, those types of constraint can be modelled as linear inequality constraints, defined as follow
  \begin{equation}
  \label{eq:contact-wrench-inequalities}
      \mathbf{C} \cdot \mathbf{f} < \mathbf{d}.
  \end{equation}
  Those constraints not only allow stable contacts at the feet, but they can also be used to regulate the grasping wrenches between the hands and the payload. In order to enforce the inequality constraints on the wrenches, the force optimization can be solved as QP problem:
  \begin{subequations}
  \begin{align}
  \label{eq:force-optimization-qp}
   \mathbf{f}^{*} = & \ \underset{\mathbf{f}}{\text{minimize}} &
    \lVert \mathbf{A} \cdot\mathbf{f} - \mathbf{b}^{*} \rVert_2, \\
    & \text{subject to} & 
    \mathbf{C} \cdot \mathbf{f} < \mathbf{d}. &
  \end{align}
  \end{subequations}
  
  \subsubsection{Task-Space Inverse Coupled Dynamics}
  The coupled-dynamics description  in Equation \eqref{eq:multi-system-equations-compact} allows to retrieve the motor torque command from a desired wrench vector $\mathbf{f}^{*}$. This process is also referred to, in the literature, as \textit{task-space inverse dynamics} \cite{delprete2015}.
  In the following formulation, however, the inverse dynamics is not solved for a single robot, but for the whole system of agents.
  The joint torques vector $\pmb{\tau}$ can be obtained as follow. From Equation \eqref{eq:multi-system-equations-compact:dynamics}, the state acceleration is retrieved as
  \begin{equation}
      \dot{\pmb{\nu}} = \mathbf{M}^{-1} \left [ \mathbf{B} \pmb{\tau} + \mathbf{J}^T \mathbf{f} - \mathbf{h} \right ] ,
  \end{equation}
  and substituting it into Equation \eqref{eq:multi-system-equations-compact:holonomics}, one obtains
  \begin{equation}
  \label{eq:dynamics-projected-into-contraints-bold}
      \mathbf{J} \cdot \mathbf{M}^{-1} \left [ \mathbf{B} \pmb{\tau} + \mathbf{J}^T \mathbf{f} - \mathbf{h} \right ] + \dot{\mathbf{J}} \ {\pmb{\nu}} = 0,
  \end{equation}
  which represents the projection of the system dynamics into the constraints. Equation \eqref{eq:dynamics-projected-into-contraints-bold} can be solved for $\pmb{\tau}$ substituting $\mathbf{f}^{*}$. Generally, a solution can be found via pseudo-inverse:
  \begin{equation}
  \pmb{\tau} = \pmb{\Lambda}^{\dagger} \left [  \mathbf{J} \cdot \mathbf{M}^{-1} (\mathbf{h} - \mathbf{J}^T \mathbf{f}) - \dot{\mathbf{J}} \ {\pmb{\nu}} \right ] + N_{\pmb{\Lambda}} \pmb{\tau}_0,
  \end{equation}
  where $\pmb{\Lambda} = \mathbf{J} \cdot \mathbf{M}^{-1} \mathbf{B} \in \mathbb{R}^{\mathbf{n}_k \times \mathbf{n}_j}$, $N_{\pmb{\Lambda}}$ is its null-space projector, and $\pmb{\tau}_0 \in \mathbb{R}^{\mathbf{n}_j}$ is a free variable. A convenient approach, adapted from literature \cite{nava2016}, consists in exploiting the free-variable to achieve a \textit{postural task}, \textit{i.e.} tracking of desired joints trajectory $(s_{(\text{R})}^{\text{des}},\dot{s}_{(\text{R})}^{\text{des}})$, choosing it as follow:
  \begin{equation}
  \label{eq:postural-task}
  \begin{aligned}
      \pmb{\tau}_0 & = \mathbf{h}_s - \mathbf{J}_s^T \mathbf{f} - \begin{bmatrix}
        {K}^{s,(\text{R})}_p & 0 \\
        0 & {K}^{s,(\text{H})}_p
      \end{bmatrix} 
      \begin{bmatrix}
        s_{(\text{R})} - s_{(\text{R})}^{\text{des}} \\ s_{(\text{H})} - s_{(\text{H})}^{\text{des}}
      \end{bmatrix} \\
      & - \begin{bmatrix}
        {K}^{s,(\text{R})}_d & 0 \\
        0 & {K}^{s,(\text{H})}_d
      \end{bmatrix}
      \begin{bmatrix}
        \dot{s}_{(\text{R})} - \dot{s}_{(\text{R})}^{\text{des}} \\ \dot{s}_{(\text{H})} - \dot{s}_{(\text{H})}^{\text{des}}
      \end{bmatrix},
  \end{aligned}
  \end{equation}
  where $\mathbf{h}_s \in \mathbb{R}^{\mathbf{n}_j}$, $\mathbf{J}_s  \in \mathbb{R}^{\mathbf{n}_k \times \mathbf{n}_j} $ are extracted from $\mathbf{h}$ and $\mathbf{J}$ by considering exclusively the joints dynamics, and ${K}^{s,(.)}_p,{K}^{s,(.)}_p \in \mathbb{R}^{n_{j,(.)} \times n_{j,(.)}}$ are positive definite matrices.

%% file: Figs/control_architecture.tikz
\begin{tikzpicture}[x=0.75pt,y=0.75pt,yscale=-1,xscale=1]

\draw   (329,130) -- (430,130) -- (430,190) -- (329,190) -- cycle ;
\draw [color={rgb, 255:red, 0; green, 0; blue, 0 }  ,draw opacity=1 ]   (171,160) -- (197,160) ;
\draw [shift={(200,160)}, rotate = 180] [fill={rgb, 255:red, 0; green, 0; blue, 0 }  ,fill opacity=1 ][line width=0.08]  [draw opacity=0] (8.93,-4.29) -- (0,0) -- (8.93,4.29) -- cycle    ;
\draw [color={rgb, 255:red, 208; green, 2; blue, 27 }  ,draw opacity=1 ]   (20.43,210.11) -- (20,159.5) -- (66,159.5) ;
\draw [shift={(69,159.5)}, rotate = 180] [fill={rgb, 255:red, 208; green, 2; blue, 27 }  ,fill opacity=1 ][line width=0.08]  [draw opacity=0] (8.93,-4.29) -- (0,0) -- (8.93,4.29) -- cycle    ;
\draw [color={rgb, 255:red, 0; green, 0; blue, 0 }  ,draw opacity=1 ]   (430,159) -- (517,160.45) ;
\draw [shift={(520,160.5)}, rotate = 180.95] [fill={rgb, 255:red, 0; green, 0; blue, 0 }  ,fill opacity=1 ][line width=0.08]  [draw opacity=0] (8.93,-4.29) -- (0,0) -- (8.93,4.29) -- cycle    ;
\draw  [fill={rgb, 255:red, 245; green, 166; blue, 35 }  ,fill opacity=0.25 ] (521.33,300) .. controls (521.33,283.16) and (534.99,269.5) .. (551.83,269.5) -- (609.5,269.5) .. controls (626.34,269.5) and (640,283.16) .. (640,300) -- (640,300) .. controls (640,316.84) and (626.34,330.5) .. (609.5,330.5) -- (551.83,330.5) .. controls (534.99,330.5) and (521.33,316.84) .. (521.33,300) -- cycle ;
\draw   (70,131) -- (171,131) -- (171,190) -- (70,190) -- cycle ;
\draw [color={rgb, 255:red, 74; green, 144; blue, 226 }  ,draw opacity=1 ]   (519.67,99.5) -- (380,100.2) -- (380,126.2) ;
\draw [shift={(380,129.2)}, rotate = 270] [fill={rgb, 255:red, 74; green, 144; blue, 226 }  ,fill opacity=1 ][line width=0.08]  [draw opacity=0] (8.93,-4.29) -- (0,0) -- (8.93,4.29) -- cycle    ;
\draw   (200,130) -- (300,130) -- (300,189) -- (200,189) -- cycle ;
\draw [color={rgb, 255:red, 0; green, 0; blue, 0 }  ,draw opacity=1 ]   (299.8,161) -- (325.67,161) ;
\draw [shift={(328.67,161)}, rotate = 180] [fill={rgb, 255:red, 0; green, 0; blue, 0 }  ,fill opacity=1 ][line width=0.08]  [draw opacity=0] (8.93,-4.29) -- (0,0) -- (8.93,4.29) -- cycle    ;
\draw [color={rgb, 255:red, 0; green, 0; blue, 0 }  ,draw opacity=1 ][line width=1.5]  [dash pattern={on 1.69pt off 2.76pt}]  (379.05,265.5) -- (379.95,196) ;
\draw [shift={(380,192)}, rotate = 90.74] [fill={rgb, 255:red, 0; green, 0; blue, 0 }  ,fill opacity=1 ][line width=0.08]  [draw opacity=0] (11.61,-5.58) -- (0,0) -- (11.61,5.58) -- cycle    ;
\draw [shift={(379,269.5)}, rotate = 270.74] [fill={rgb, 255:red, 0; green, 0; blue, 0 }  ,fill opacity=1 ][line width=0.08]  [draw opacity=0] (11.61,-5.58) -- (0,0) -- (11.61,5.58) -- cycle    ;
\draw [color={rgb, 255:red, 0; green, 0; blue, 0 }  ,draw opacity=1 ][line width=1.5]  [dash pattern={on 1.69pt off 2.76pt}]  (120,266.5) -- (120,194.5) ;
\draw [shift={(120,190.5)}, rotate = 90] [fill={rgb, 255:red, 0; green, 0; blue, 0 }  ,fill opacity=1 ][line width=0.08]  [draw opacity=0] (11.61,-5.58) -- (0,0) -- (11.61,5.58) -- cycle    ;
\draw [shift={(120,270.5)}, rotate = 270] [fill={rgb, 255:red, 0; green, 0; blue, 0 }  ,fill opacity=1 ][line width=0.08]  [draw opacity=0] (11.61,-5.58) -- (0,0) -- (11.61,5.58) -- cycle    ;
\draw  [color={rgb, 255:red, 208; green, 2; blue, 27 }  ,draw opacity=1 ][dash pattern={on 4.5pt off 4.5pt}] (0.63,229.91) .. controls (0.63,218.98) and (9.49,210.11) .. (20.43,210.11) .. controls (31.36,210.11) and (40.23,218.98) .. (40.23,229.91) .. controls (40.23,240.85) and (31.36,249.71) .. (20.43,249.71) .. controls (9.49,249.71) and (0.63,240.85) .. (0.63,229.91) -- cycle ;
\draw [color={rgb, 255:red, 208; green, 2; blue, 27 }  ,draw opacity=1 ]   (20.43,249.71) -- (20,298.5) -- (67,299.44) ;
\draw [shift={(70,299.5)}, rotate = 181.15] [fill={rgb, 255:red, 208; green, 2; blue, 27 }  ,fill opacity=1 ][line width=0.08]  [draw opacity=0] (8.93,-4.29) -- (0,0) -- (8.93,4.29) -- cycle    ;
\draw  [dash pattern={on 4.5pt off 4.5pt}] (500,287.84) .. controls (500,272.46) and (512.46,260) .. (527.84,260) -- (631.16,260) .. controls (646.54,260) and (659,272.46) .. (659,287.84) -- (659,371.36) .. controls (659,386.74) and (646.54,399.2) .. (631.16,399.2) -- (527.84,399.2) .. controls (512.46,399.2) and (500,386.74) .. (500,371.36) -- cycle ;
\draw  [dash pattern={on 4.5pt off 4.5pt}] (501,87.12) .. controls (501,71.59) and (513.59,59) .. (529.12,59) -- (630.88,59) .. controls (646.41,59) and (659,71.59) .. (659,87.12) -- (659,171.48) .. controls (659,187.01) and (646.41,199.6) .. (630.88,199.6) -- (529.12,199.6) .. controls (513.59,199.6) and (501,187.01) .. (501,171.48) -- cycle ;
\draw  [fill={rgb, 255:red, 74; green, 144; blue, 226 }  ,fill opacity=0.25 ] (519.67,99.5) .. controls (519.67,82.66) and (533.32,69) .. (550.17,69) -- (608.5,69) .. controls (625.34,69) and (639,82.66) .. (639,99.5) -- (639,99.5) .. controls (639,116.34) and (625.34,130) .. (608.5,130) -- (550.17,130) .. controls (533.32,130) and (519.67,116.34) .. (519.67,99.5) -- cycle ;
\draw  [fill={rgb, 255:red, 245; green, 166; blue, 35 }  ,fill opacity=0.25 ] (520,160.5) .. controls (520,143.66) and (533.66,130) .. (550.5,130) -- (609.5,130) .. controls (626.34,130) and (640,143.66) .. (640,160.5) -- (640,160.5) .. controls (640,177.34) and (626.34,191) .. (609.5,191) -- (550.5,191) .. controls (533.66,191) and (520,177.34) .. (520,160.5) -- cycle ;
\draw [color={rgb, 255:red, 208; green, 2; blue, 27 }  ,draw opacity=1 ]   (580,255.6) -- (580,203.6) ;
\draw [shift={(580,200.6)}, rotate = 90] [fill={rgb, 255:red, 208; green, 2; blue, 27 }  ,fill opacity=1 ][line width=0.08]  [draw opacity=0] (8.93,-4.29) -- (0,0) -- (8.93,4.29) -- cycle    ;
\draw [shift={(580,258.6)}, rotate = 270] [fill={rgb, 255:red, 208; green, 2; blue, 27 }  ,fill opacity=1 ][line width=0.08]  [draw opacity=0] (8.93,-4.29) -- (0,0) -- (8.93,4.29) -- cycle    ;
\draw   (329,269) -- (430,269) -- (430,329) -- (329,329) -- cycle ;
\draw   (70,270) -- (171,270) -- (171,329) -- (70,329) -- cycle ;
\draw [color={rgb, 255:red, 74; green, 144; blue, 226 }  ,draw opacity=1 ]   (380,100.2) -- (250,100.2) -- (250.9,127.2) ;
\draw [shift={(251,130.2)}, rotate = 268.09] [fill={rgb, 255:red, 74; green, 144; blue, 226 }  ,fill opacity=1 ][line width=0.08]  [draw opacity=0] (8.93,-4.29) -- (0,0) -- (8.93,4.29) -- cycle    ;
\draw   (201,330) -- (301,330) -- (301,389) -- (201,389) -- cycle ;
\draw [color={rgb, 255:red, 65; green, 117; blue, 5 }  ,draw opacity=1 ]   (251,329.2) -- (250.02,192.8) ;
\draw [shift={(250,189.8)}, rotate = 89.59] [fill={rgb, 255:red, 65; green, 117; blue, 5 }  ,fill opacity=1 ][line width=0.08]  [draw opacity=0] (8.93,-4.29) -- (0,0) -- (8.93,4.29) -- cycle    ;
\draw [color={rgb, 255:red, 65; green, 117; blue, 5 }  ,draw opacity=1 ]   (520,360) -- (306,359.21) ;
\draw [shift={(303,359.2)}, rotate = 0.21] [fill={rgb, 255:red, 65; green, 117; blue, 5 }  ,fill opacity=1 ][line width=0.08]  [draw opacity=0] (8.93,-4.29) -- (0,0) -- (8.93,4.29) -- cycle    ;
\draw  [fill={rgb, 255:red, 65; green, 117; blue, 5 }  ,fill opacity=0.25 ] (520,360) .. controls (520,343.71) and (533.21,330.5) .. (549.5,330.5) -- (610.5,330.5) .. controls (626.79,330.5) and (640,343.71) .. (640,360) -- (640,360) .. controls (640,376.29) and (626.79,389.5) .. (610.5,389.5) -- (549.5,389.5) .. controls (533.21,389.5) and (520,376.29) .. (520,360) -- cycle ;
\draw [color={rgb, 255:red, 65; green, 117; blue, 5 }  ,draw opacity=1 ]   (250,299.2) -- (325,300.16) ;
\draw [shift={(328,300.2)}, rotate = 180.73] [fill={rgb, 255:red, 65; green, 117; blue, 5 }  ,fill opacity=1 ][line width=0.08]  [draw opacity=0] (8.93,-4.29) -- (0,0) -- (8.93,4.29) -- cycle    ;
\draw  [color={rgb, 255:red, 65; green, 117; blue, 5 }  ,draw opacity=1 ][dash pattern={on 6.75pt off 4.5pt}][line width=2.25]  (48,279.44) .. controls (48,261.52) and (62.52,247) .. (80.44,247) -- (425.56,247) .. controls (443.48,247) and (458,261.52) .. (458,279.44) -- (458,376.76) .. controls (458,394.68) and (443.48,409.2) .. (425.56,409.2) -- (80.44,409.2) .. controls (62.52,409.2) and (48,394.68) .. (48,376.76) -- cycle ;

\draw (82,145) node [anchor=north west][inner sep=0.75pt]   [align=left] {\begin{minipage}[lt]{47.69pt}\setlength\topsep{0pt}
\begin{center}
Trajectory\\Planning
\end{center}

\end{minipage}};
\draw (353,145) node [anchor=north west][inner sep=0.75pt]   [align=left] {\begin{minipage}[lt]{35.6pt}\setlength\topsep{0pt}
\begin{center}
Trajectory \\Control
\end{center}

\end{minipage}};
\draw (549,283.5) node [anchor=north west][inner sep=0.75pt]   [align=left] {\begin{minipage}[lt]{45.81pt}\setlength\topsep{0pt}
\begin{center}
Muscle \\Activation
\end{center}

\end{minipage}};
\draw (208,145) node [anchor=north west][inner sep=0.75pt]   [align=left] {\begin{minipage}[lt]{53.75pt}\setlength\topsep{0pt}
\begin{center}
Trajectory\\Adjustment
\end{center}

\end{minipage}};
\draw (3,222.4) node [anchor=north west][inner sep=0.75pt]  [color={rgb, 255:red, 208; green, 2; blue, 27 }  ,opacity=1 ] [align=left] {\begin{minipage}[lt]{23.69pt}\setlength\topsep{0pt}
\begin{center}
Task
\end{center}

\end{minipage}};
\draw (588,37) node [anchor=north west][inner sep=0.75pt]   [align=left] {\begin{minipage}[lt]{48.16pt}\setlength\topsep{0pt}
\begin{center}
Robot $\displaystyle ( R)$
\end{center}

\end{minipage}};
\draw (550,84) node [anchor=north west][inner sep=0.75pt]   [align=left] {\begin{minipage}[lt]{42.98pt}\setlength\topsep{0pt}
\begin{center}
Sensory \\System
\end{center}

\end{minipage}};
\draw (546.33,144) node [anchor=north west][inner sep=0.75pt]   [align=left] {\begin{minipage}[lt]{45.81pt}\setlength\topsep{0pt}
\begin{center}
Motor \\Actuators
\end{center}

\end{minipage}};
\draw (577,402) node [anchor=north west][inner sep=0.75pt]   [align=left] {\begin{minipage}[lt]{55.11pt}\setlength\topsep{0pt}
\begin{center}
Human $\displaystyle ( H)$
\end{center}

\end{minipage}};
\draw (588,210) node [anchor=north west][inner sep=0.75pt]   [align=left] {\begin{minipage}[lt]{45.59pt}\setlength\topsep{0pt}
\begin{center}
{\small Physical}\\{\small Interaction}
\end{center}

\end{minipage}};
\draw (81,284) node [anchor=north west][inner sep=0.75pt]   [align=left] {\begin{minipage}[lt]{47.69pt}\setlength\topsep{0pt}
\begin{center}
Trajectory\\Planning
\end{center}

\end{minipage}};
\draw (353,284) node [anchor=north west][inner sep=0.75pt]   [align=left] {\begin{minipage}[lt]{35.6pt}\setlength\topsep{0pt}
\begin{center}
Motor \\Control
\end{center}

\end{minipage}};
\draw (215,344) node [anchor=north west][inner sep=0.75pt]   [align=left] {\begin{minipage}[lt]{51.48pt}\setlength\topsep{0pt}
\begin{center}
Human\\Perception
\end{center}

\end{minipage}};
\draw (550,341.5) node [anchor=north west][inner sep=0.75pt]   [align=left] {\begin{minipage}[lt]{46.19pt}\setlength\topsep{0pt}
\begin{center}
Wearable\\Sensors
\end{center}

\end{minipage}};
\draw (51,416) node [anchor=north west][inner sep=0.75pt]   [align=left] {\begin{minipage}[lt]{150.61pt}\setlength\topsep{0pt}
\begin{center}
\textcolor[rgb]{0.25,0.46,0.02}{Human Behavioural Model}
\end{center}

\end{minipage}};
\draw (456,139) node [anchor=north west][inner sep=0.75pt]   [align=left] {\begin{minipage}[lt]{15.15pt}\setlength\topsep{0pt}
\begin{center}
$\displaystyle \tau _{R}$
\end{center}

\end{minipage}};
\draw (271,79) node [anchor=north west][inner sep=0.75pt]   [align=left] {\begin{minipage}[lt]{31.86pt}\setlength\topsep{0pt}
\begin{center}
$\displaystyle q_{R} ,\nu _{R}$
\end{center}

\end{minipage}};
\draw (225.59,312.54) node [anchor=north west][inner sep=0.75pt]  [rotate=-269.79] [align=left] {\begin{minipage}[lt]{33.66pt}\setlength\topsep{0pt}
\begin{center}
$\displaystyle q_{H} ,\nu _{H}$
\end{center}

\end{minipage}};

\end{tikzpicture}

%% file: sections/experimental_validation.tex
\section{Experimental Validation}
\label{sec:experimental-validation}

The proposed framework is implemented and validated in an experimental scenario where a humanoid robot and a sensorized human subject perform a payload lifting task.

\subsection{Experimental Setup}
The control infrastructure has been tested using the iCub3 humanoid robot platform \cite{dafarra2022}, which, for the purpose, was endowed with 22 degrees of freedom. 
The human subject is tracked using the iFeel wearable suit, and is modelled as a 29 degrees of freedom multi-body system. 
According to the capability of the robot, a payload of $\SI{4}{\kilogram}$ is used. 
The experimental setup is shown in Figure \ref{fig:human-robot-collaborative-lifting}.
    
\subsection{Implementation Details}
\label{sec:experimental_validatoin_implementation}
The force optimization and task-space inverse coupled dynamics, introduced in Section \ref{sec:motor-control-robot}, are implemented as a single QP optimization allowing to enforce a force ergonomics minimizing the energy expenditure \cite{rapetti2021shared}. This is possible by defining a human motor control model that relates the torques generated by the two agents.

Some of the quantities from the human momentum control model in Equation \eqref{eq:momentum-control-human}, are not trivial to be determined, and  some simplifications are required.
The momentum feedback term $K^{(\text{H})}_d({L}_{(\text{H})}-{L}_{(\text{H})}^{\text{des}})$ appearing in Equation \eqref{eq:object-control-human-robot} is removed, and the estimated human momentum derivative replace the desired one, \textit{i.e.} ${\dot{{L}}}_{(\text{H})}^{\text{des}}=\hat{\dot{{L}}}_{(\text{H})}$.
The desired object height is computed from the human hands position which acts as master (human hands have an similar role in the trajectory advancement strategy discussed in Section \ref{sec:trajectory-planning-and-advancement}), while its desired orientation is maintained constant. Moreover, a gravity compensation task is used as regularization for the human internal joint torque solution, such that
    \begin{equation}
        \tau_{0,(\text{H})} = h_{s,(\text{H})} - J_{s,\text{H}}^T \mathrm{f}_{(\text{H})}.
    \end{equation}

Combining the human motor control model, the force ergonomics optimization, and adding an inverse dynamics task, the following hierarchical optimization is implemented
\begin{subequations}
  \label{eq:human-robot-collaborative-lifting-optimization}
  \begin{align}
   \mathbf{f}^{*} = &  \underset{\mathbf{f}}{\text{argmin }} &
    \lVert\pmb{\tau}^{*}(\mathbf{f}) \rVert_{W_{\pmb{\tau}}} &  \notag \\
    & & + \ \lVert \mathrm{f}_{\mathcal{K},(\text{H})} - \hat{\mathrm{f}}_{\mathcal{K},(\text{H})} \rVert_{W_{\hat{\mathrm{f}}}}, &    
    \label{eq:hrclfo:torque-minimization} \\
    & \text{s.t.: } & \mathbf{b}^{*} = \mathbf{A} \cdot \mathbf{f}, &  
    \label{eq:hrclfo:force-tasks}\\
    & & \mathbf{C} \cdot \mathbf{f} < \mathbf{d}, &  
    \label{eq:hrclfo:inequality}
    \\
    & \pmb{\tau}^{*}(\mathbf{f}) = \underset{\pmb{\tau}}{\text{argmin }} & \lVert \pmb{\tau}(\mathbf{f}) - \pmb{\tau}_{0}\rVert , & \\
    &  \text{ \ \ \ \ \ \ \ \ \ \ s.t.: } & \mathbf{M} \dot{\pmb{\nu}} + \mathbf{h} = \mathbf{B} \pmb{\tau} + \mathbf{J}^T \mathbf{f}, 
    \label{eq:hrclfo:dynamics} \\
    & & \mathbf{J} \dot{\pmb{\nu}} + \dot{\mathbf{J}} {\pmb{\nu}} = 0,
    \label{eq:hrclfo:holonomic} \\
    & & \tau_{0,(\text{R})} = h_{s,(\text{R})} - J_{s,\text{R}}^T \mathrm{f}_{(\text{R})} \notag \\
    & & \ \ \ - K_{p,(\text{R})}^s (s_{(\text{R})} - s_{(\text{R})}^{\text{des}}) \notag \\
    & & \ \ \ - {K}_{d,(\text{R})}^s (\dot{{s}}_{(\text{R})} - \dot{{s}}_{(\text{R})}^{\text{des}}), 
    \label{eq:hrclfo:postural-task-robot} \\
    \label{eq:hrclfo:postural-task-human}
    & & \tau_{0,(\text{H})} = h_{s,(\text{H})} - J_{s,(\text{H})}^T \mathrm{f}_{(\text{H})},
 \end{align}
\end{subequations}
where:
\begin{itemize}
      \item Equation \eqref{eq:hrclfo:torque-minimization} applies force ergonomics minimizing the energy expenditure via torques minimization, and applies regularization on human feet forces using shoes wrench measurement denoted as $\hat{\mathrm{f}}_{\mathcal{K},(\text{H})}$ ;
      \item Equation \eqref{eq:hrclfo:force-tasks} corresponds to Equation \eqref{eq:compact-force-tasks}, and corresponds to force optimizatio tasks;
      \item Equation \eqref{eq:hrclfo:inequality} applies the inequality constraints contact wrenches to ensure unilaterality, static friction, center of pressure, and grasping feasibility;
      \item Equation \eqref{eq:hrclfo:dynamics} and \eqref{eq:hrclfo:holonomic} define the coupled dynamics formulation presented in Section \ref{sec:coupled-dynamics};
      \item Equation \eqref{eq:hrclfo:postural-task-robot} regulates the robot torques free variable, while Equation \eqref{eq:hrclfo:postural-task-human} defines the gravity compensation model for the human.
\end{itemize}

\subsection{Results}
The proposed controller implementation allows to perform collaborative lifting task in the proposed scenario. The code and video are available at {\small \url{https://github.com/ami-iit/paper_rapetti_2023_icra_eronomic_payload_lifting}}. 

\begin{figure}
  \centering
  \begin{subfigure}{0.3\columnwidth}
    \centering
    \includegraphics[trim={0 1.75cm 0 1.75cm},clip,width=\columnwidth]{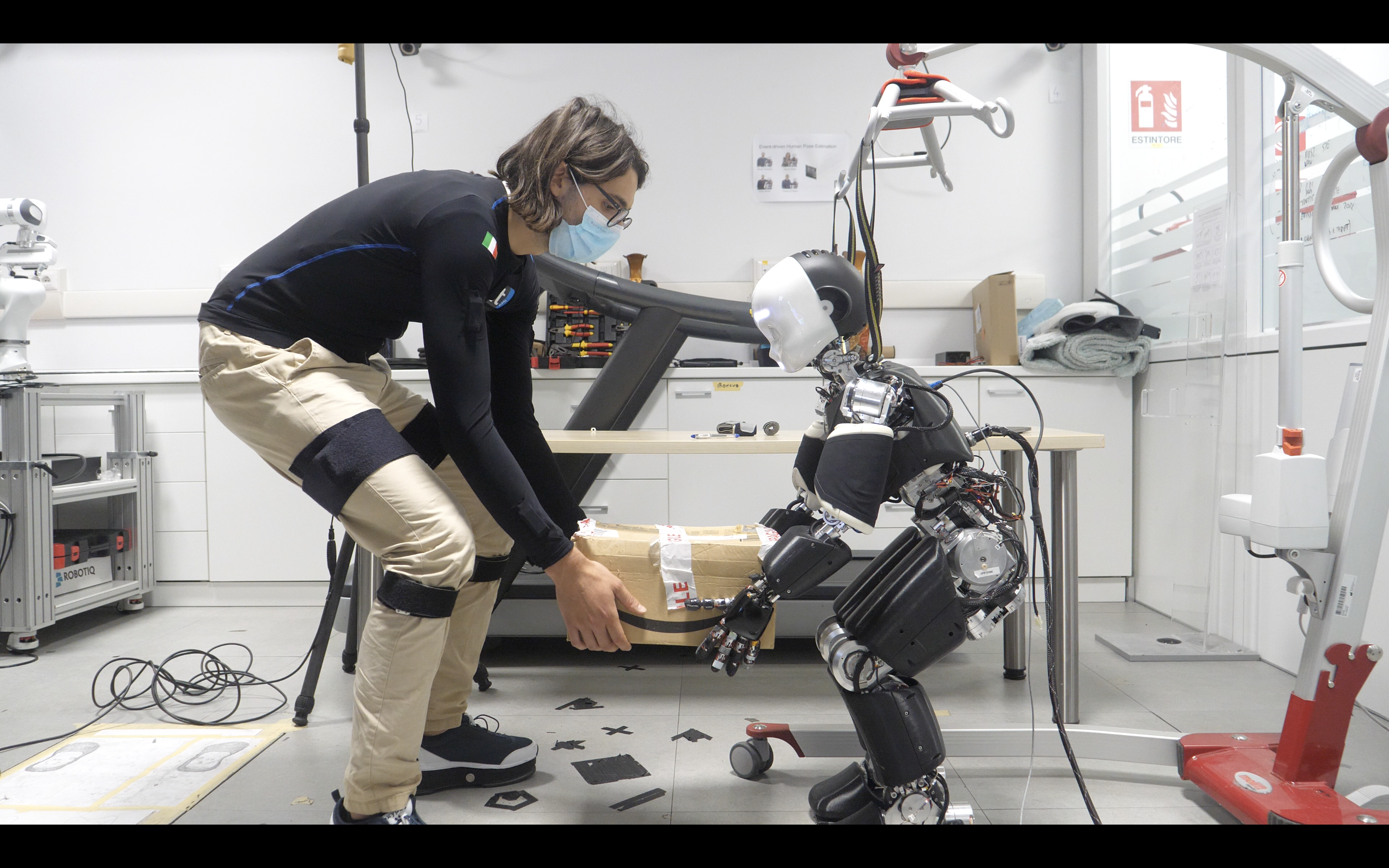}
    \caption{}
    \label{fig:collaborative-lifting-results-side:lifting-real-1}
    \end{subfigure} 
   \begin{subfigure}{0.3\columnwidth}
    \centering
    \includegraphics[trim={0 1.75cm 0 1.75cm},clip,width=\columnwidth]{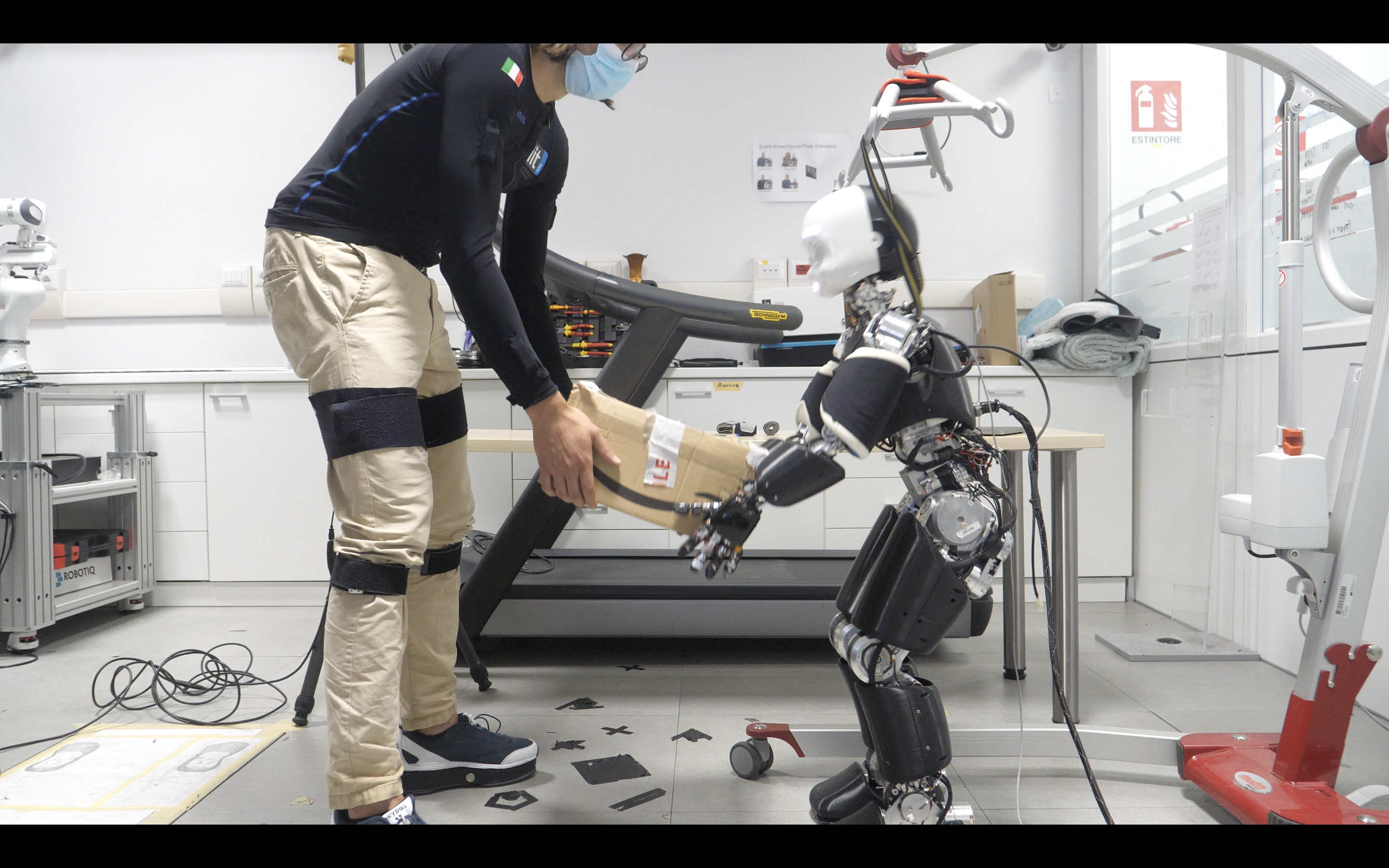}
    \caption{}
    \label{fig:collaborative-lifting-results-side:lifting-real-2}
    \end{subfigure} 
   \begin{subfigure}{0.3\columnwidth}
    \centering
    \includegraphics[trim={0 1.75cm 0 1.75cm},clip,width=\columnwidth]{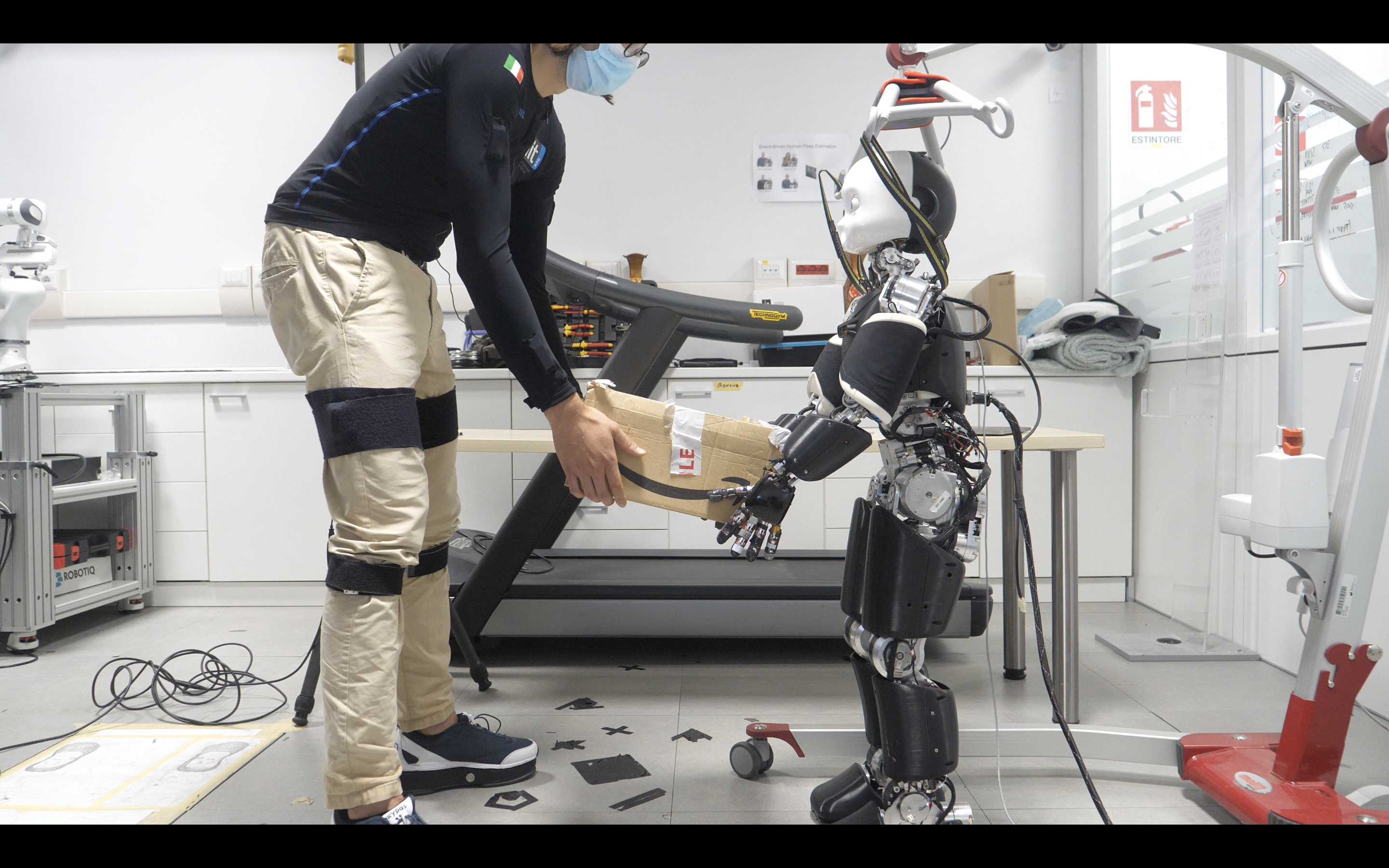}
    \caption{}
    \label{fig:collaborative-lifting-results-side:lifting-real-3}
    \end{subfigure} 
    
    \vspace{0.2cm}
  \begin{subfigure}{0.3\columnwidth}
    \centering
    \includegraphics[width=\columnwidth]{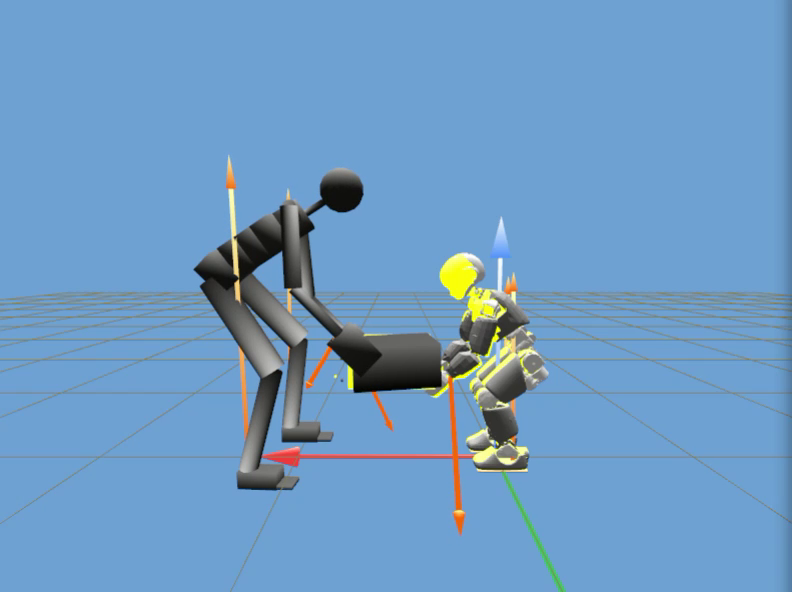}
    \caption{}
    \label{fig:collaborative-lifting-results-side:lifting-1}
    \end{subfigure} 
  \begin{subfigure}{0.3\columnwidth}
    \centering
      \includegraphics[width=\columnwidth]{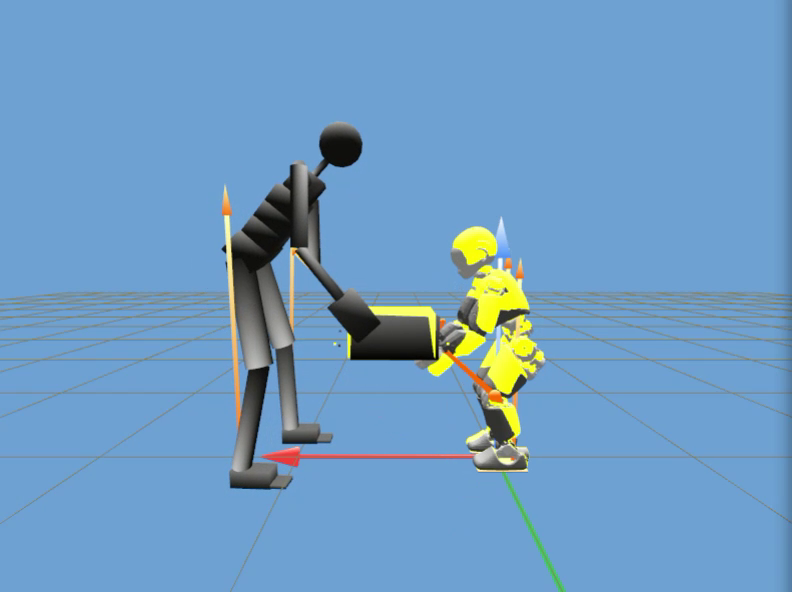}
    \caption{}
    \label{fig:collaborative-lifting-results-side:lifting-2}
    \end{subfigure}
  \begin{subfigure}{0.3\columnwidth}
    \centering
      \includegraphics[width=\columnwidth]{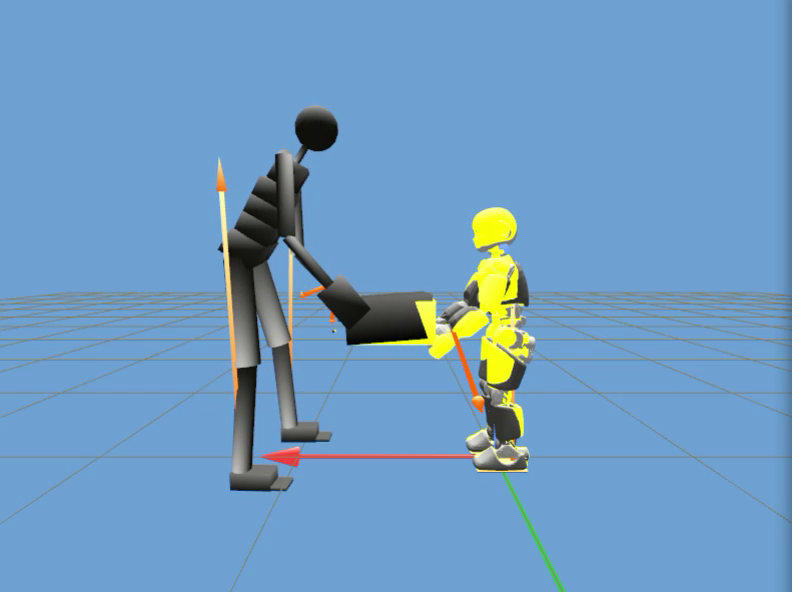}
    \caption{}
    \label{fig:collaborative-lifting-results-side:lifting-3}
    \end{subfigure}
  
  \vspace{0.2cm}
  
  \begin{subfigure}{\columnwidth}
    \centering
    \includegraphics[trim=3cm 0cm 3cm 0cm, clip=true, width=\columnwidth]{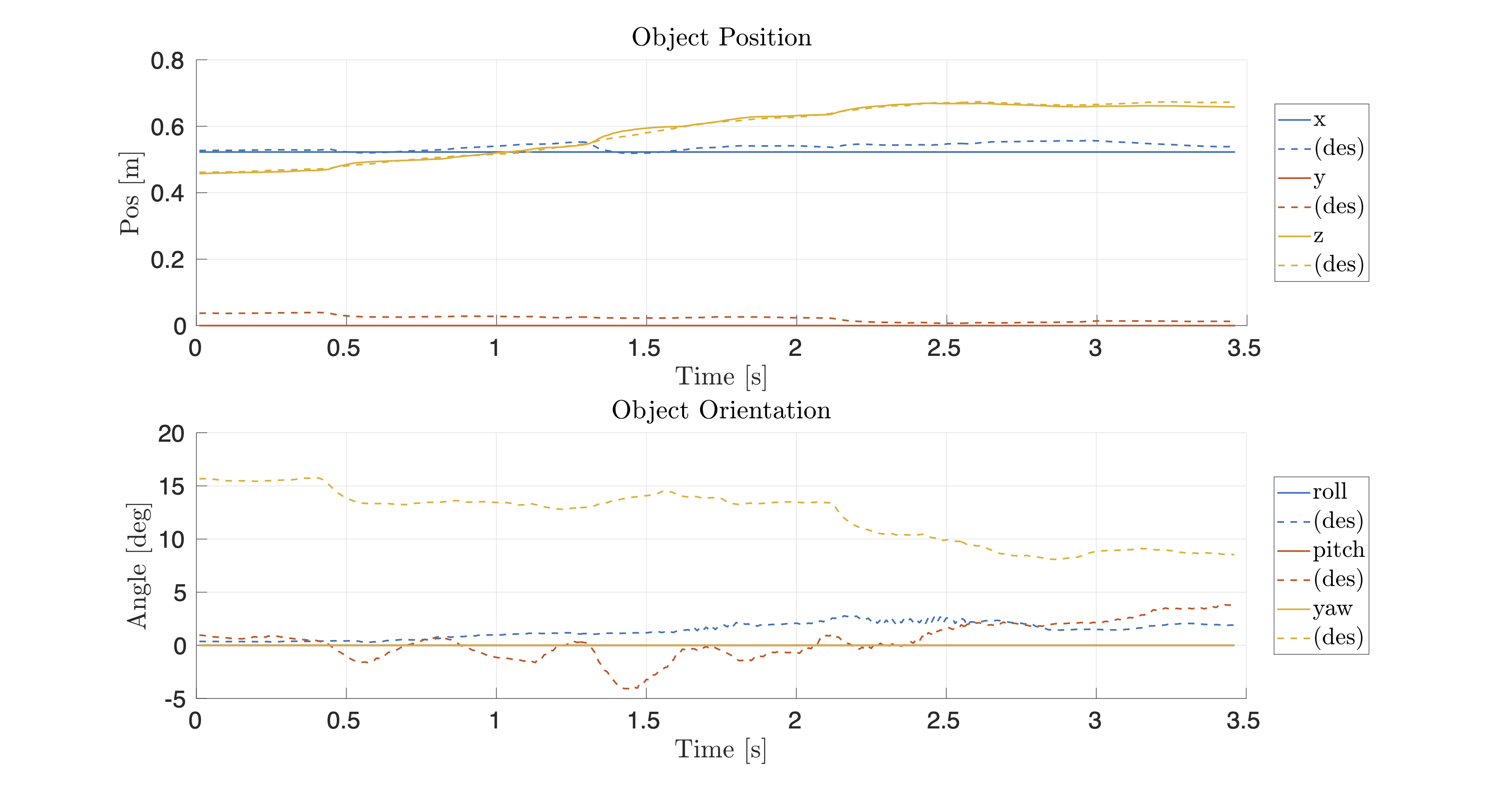}
    \caption{}
    \end{subfigure} 
    
    \vspace{0.2cm}
    
  \begin{subfigure}{\columnwidth}
    \centering
    \includegraphics[trim=3cm 0cm 3cm 0cm, clip=true,width=\columnwidth]{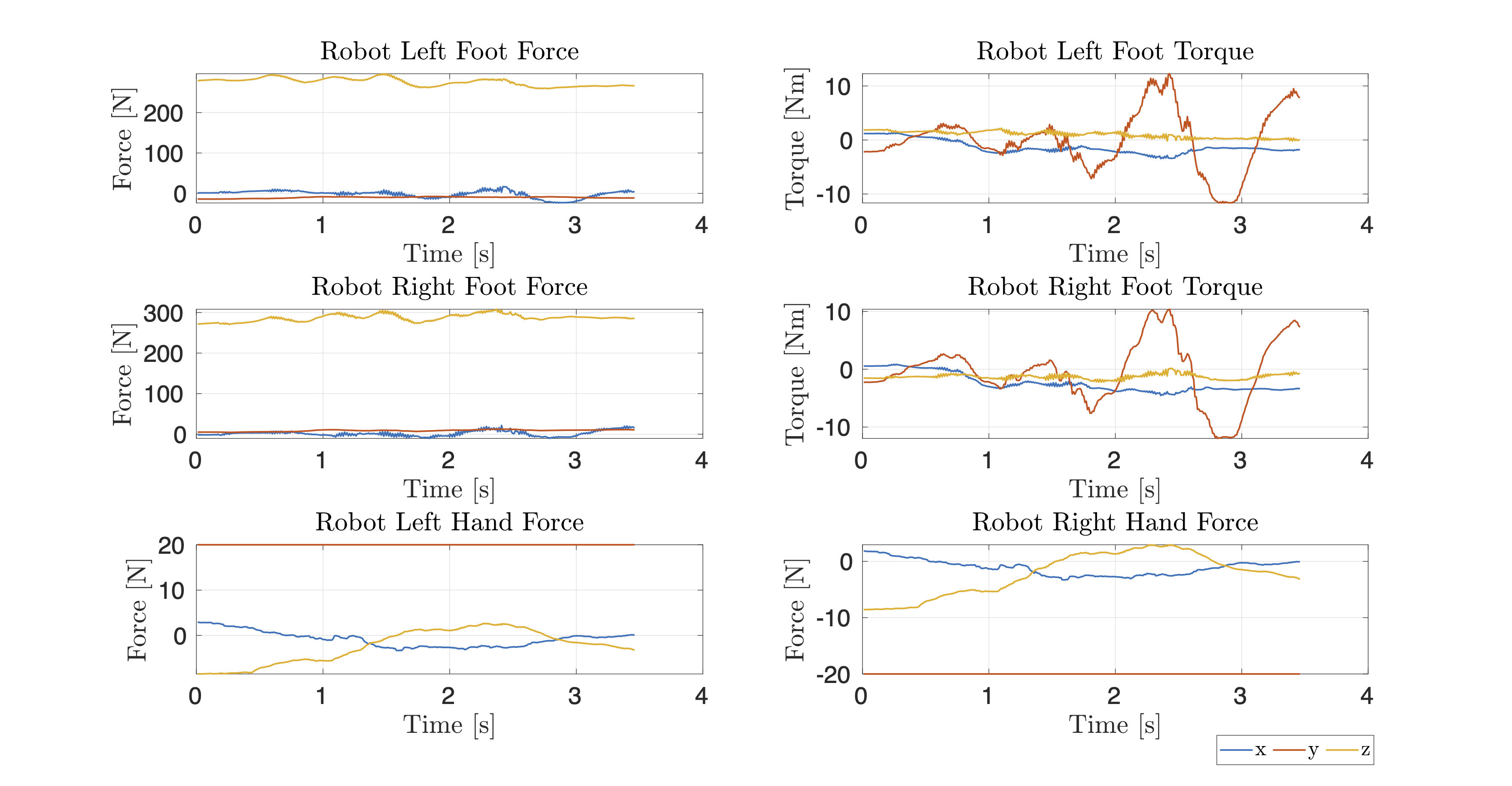}
    \end{subfigure} 
  \begin{subfigure}{\columnwidth}
    \centering
    \includegraphics[trim=3cm 0cm 3cm 0cm, clip=true,width=\columnwidth]{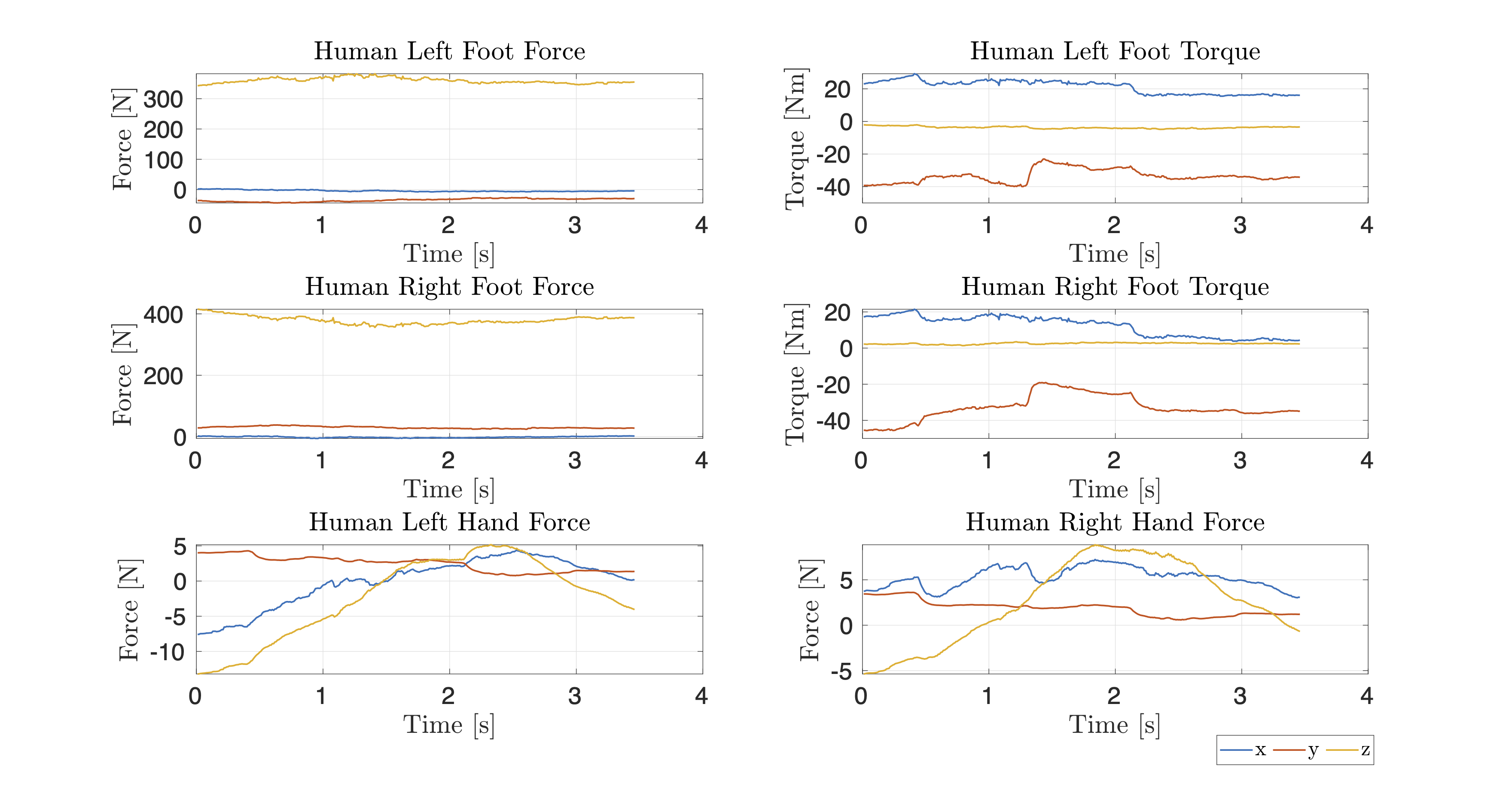}
    \caption{}
    \label{fig:collaborative-lifting-results-side:wrenches}
    \end{subfigure}
  \caption[Human-robot collaborative lifting results, fast lifting.]{Human-robot collaborative lifting experiments. Pictures \textit{(a)}-\textit{(c)} are captured during the execution, while \textit{(d)}-\textit{(f)} represents the measured configuration of human, robot, and payload in \textit{grey}, and in \textit{yellow} the reference configuration of robot and payload. \textit{(g)} and \textit{(h)} show respectively the object pose tracking performances and the computed wrenches $\textbf{f}^{*}$.}
      \label{fig:collaborative-lifting-results-side}
\end{figure}

\begin{figure}[b]
  \centering
      
  \begin{subfigure}{0.28\columnwidth}
    \centering
      \includegraphics[width=\columnwidth]{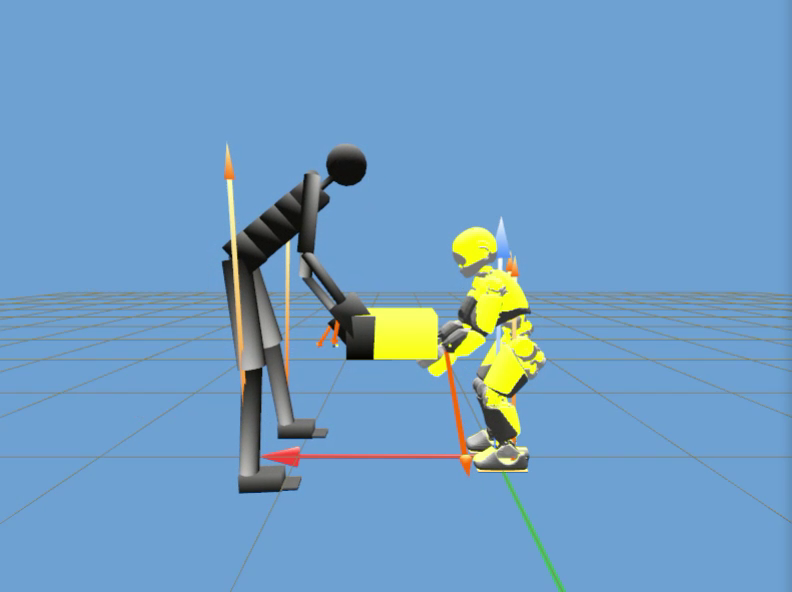}
    \caption{} 
    \end{subfigure}
  \begin{subfigure}{0.28\columnwidth}
    \centering
      \includegraphics[width=\columnwidth]{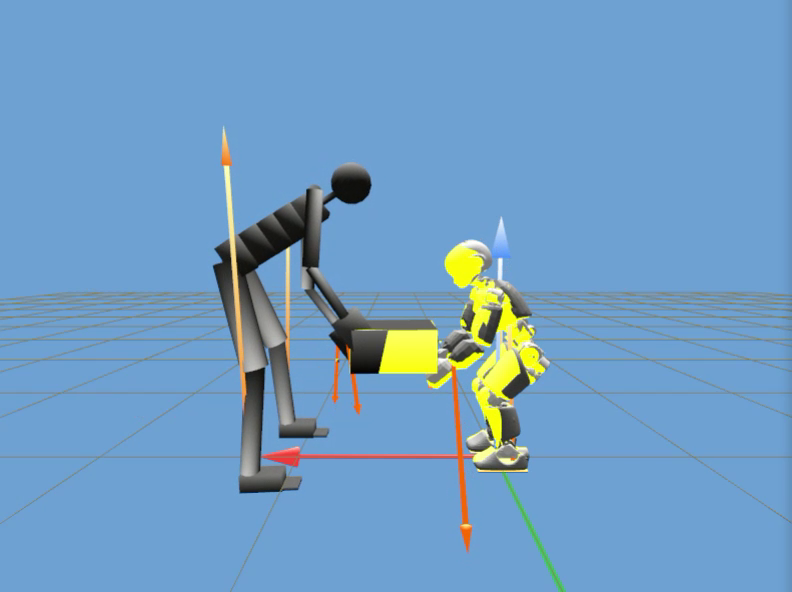}
    \caption{} 
    \end{subfigure}
  \begin{subfigure}{0.28\columnwidth}
    \centering
      \includegraphics[width=\columnwidth]{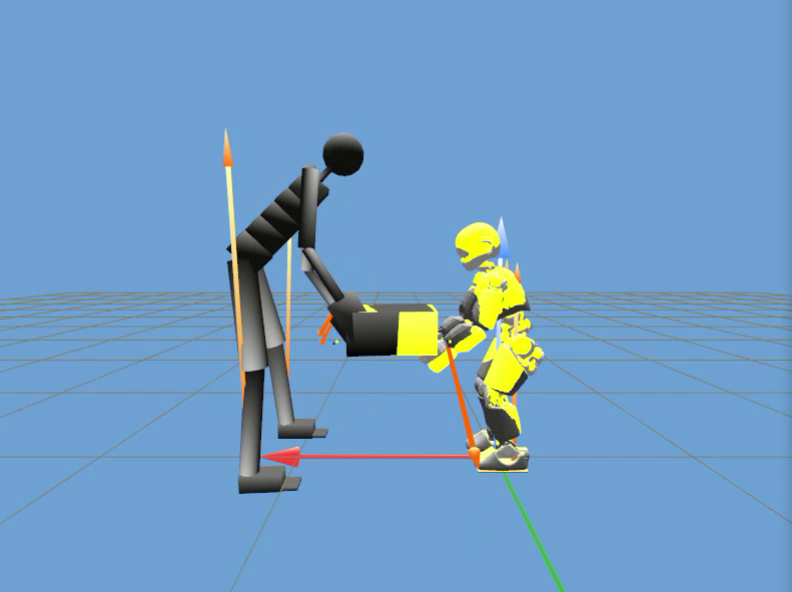}
    \caption{} 
    \end{subfigure}
  \begin{subfigure}{0.94\columnwidth}
    \centering
    \includegraphics[trim=3cm 0cm 3cm 0cm, clip=true,width=\columnwidth]{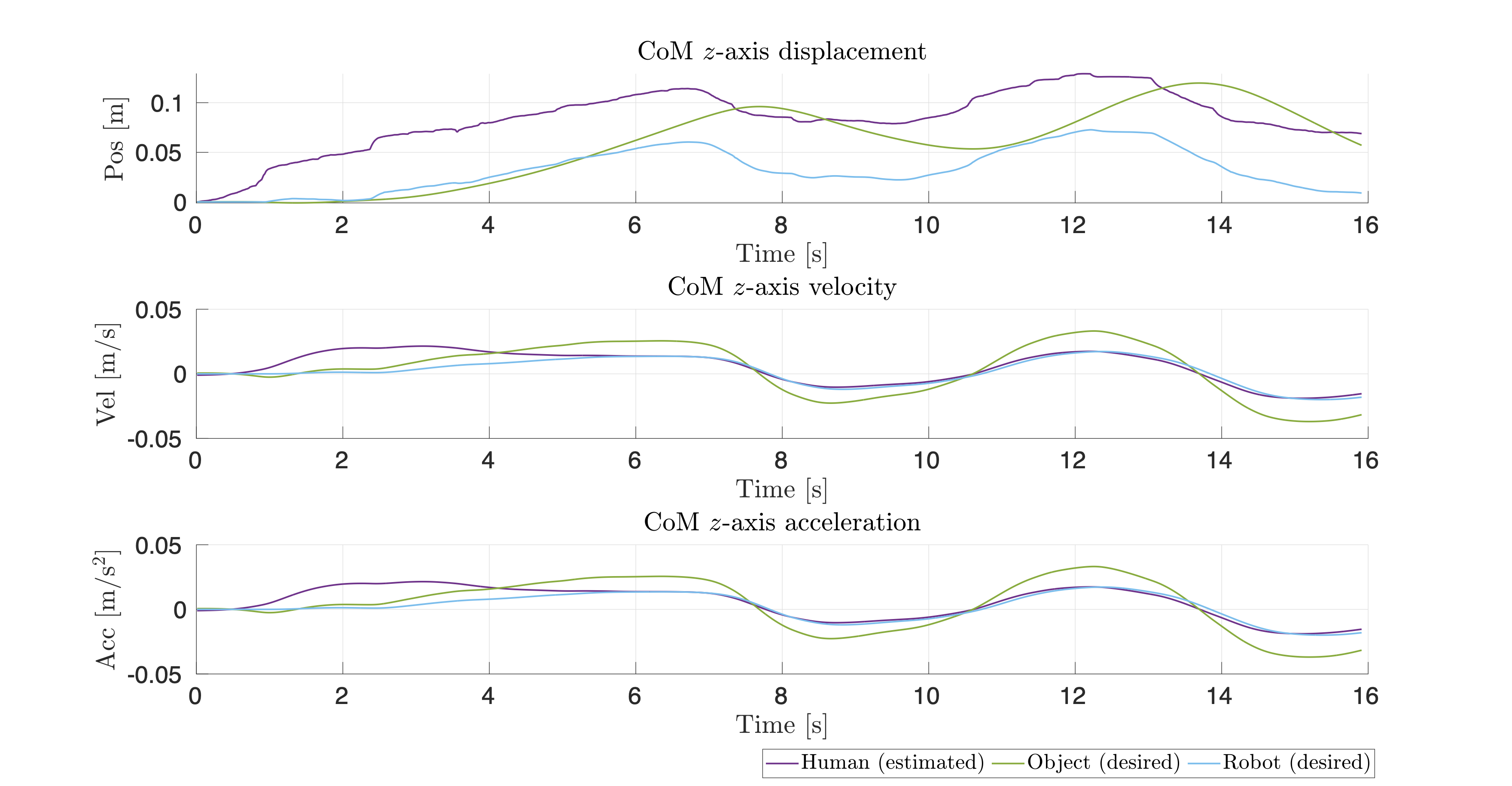}
    \caption{}
    \end{subfigure} 
  \caption[Human-Robot collaborative controller trajectory advancement.]{Trajectory advancement in collaborative controller. \textit{(a)}-\textit{(c)} display motion sequence: object is moved upwards, \textit{(a)} and \textit{(c)}, and downwards, \textit{(b)}. \textit{(d)} shows the estimated and reference trajectories for the CoM $z$-axis.}
      \label{fig:collaborative-lifting-up-down}
\end{figure}

Robot configurations at height $\SI{35}{\cm}$ and $\SI{75}{\cm}$ are computed. The transition between the configurations is handled by the trajectory advancement strategy proposed in Section \ref{sec:trajectory-planning-and-advancement}. Figure \ref{fig:collaborative-lifting-results-side} presents the results for a lifting tasks experiment. The robot grasp the payload from the side, and regulates the wrenches following the motion of the human collaborator. Figure \ref{fig:collaborative-lifting-results-side:lifting-1}-\ref{fig:collaborative-lifting-results-side:lifting-3} represents the data collected during the lifting motion, showing the measured configuration of human, robot, and payload, and the desired configuration of the robot and payload, in yellow.
The plots presents the tracking performances for the object position and orientation. Comparable performances are observed for the three position components, which retains the error $< \SI{5}{\cm}$. Compared to the other axis, yaw tracking presents the worst performances. Since the estimation of the object pose relies on the kinematic of the human, errors might be affected by modelling error. Concerning the desired wrenches, it can be observed from Figure \ref{fig:collaborative-lifting-results-side:wrenches} that a fixed compression wrench of $\SI{20}{\newton}$, determined empirically, has been enforced as constraint in order to maintain the grasp on the object.
Figure \ref{fig:collaborative-lifting-up-down} shows a second experiment.
In this case the subject moves the payload up and down. As a consequence, the robot reference trajectories follow the human motion along the trajectory.
It is important to note that both experiments use the same controller, and the speed and direction of the robot movement are automatically adjusted based on the human motion.

%% file: sections/conclusions.tex
\section{Conclusions}
\label{sec:conclusions}
This paper proposes a framework that allows a humanoid robot  to perform collaborative payload lifting task with a human partner. Using wearable sensors and coupled dynamics, the controller is designed to optimize the task execution
taking into consideration the ergonomics
of both agents through symmetrical models for humans and robots. Moreover, the robot has the capability to adapt its trajectory to follow human motion. The implementation of the framework has been tested on robot hardware.
Since a single scenario has been tested, extended validation should be performed with a group of operators, collecting feedback on the quality of the lifting.
Adding more sophisticated human behavioural models, such as black-box and data-driven, can enrich the framework towards predictive control strategies while introducing less-invasive monitoring systems can facilitate transferring the proposed method to real-world scenarios.